\documentclass[journal]{IEEEtran}

\usepackage{amsmath}
\usepackage{amssymb}
\usepackage{array}
\usepackage{booktabs}
\usepackage{graphicx}
\usepackage{multirow}
\usepackage{pdflscape}
\usepackage{rotating}
\usepackage{siunitx}
\usepackage{tabularx}
\usepackage{textcomp}
\usepackage{url}
\usepackage{cite}
\usepackage{algorithm}
\usepackage{algpseudocode}
\usepackage{soul}
\renewcommand{\hl}[1]{#1}
\usepackage{xurl}
\usepackage{hyperref}
\soulregister\cite7
\soulregister\ref7
\usepackage[switch]{lineno}

\algrenewcommand\algorithmicrequire{\textbf{Inputs:}}
\algrenewcommand\algorithmicensure{\textbf{Outputs:}}

\begin{document}

\title{Retrieval of Coastal \hl{Water Quality} Parameters From Near-Surface Hyperspectral Remote Sensing Reflectance Using Physics-Aware Meta-Learning}

\author{Yiqing~Guo,
Nagur~R.~C.~Cherukuru,
Eric~A.~Lehmann,
S.~L.~Kesav~Unnithan,
Tim~J.~Malthus,
Gemma~Kerrisk,
Xiubin~Qi,
Faisal~Islam,
Tisham~Dhar,
and~Mark~J.~Doubell%
\thanks{
This work is supported by Commonwealth Scientific and Industrial Research Organisation (CSIRO) AquaWatch Australia Program, CSIRO AI4Missions, South Australian Research and Development Institute (SARDI), CSIRO Technology, CSIRO Environment, CSIRO Space and Astronomy, CSIRO Earth Analytics Science and Innovation (EASI) platform, and CSIRO AquaWatch Data Service (ADS). \textit{(Corresponding author: Yiqing Guo.)}}
\thanks{Yiqing Guo and Eric A. Lehmann are with CSIRO Technology, Acton, ACT 2601, Australia (e-mail: yiqing.guo@csiro.au; eric.lehmann@csiro.au).}%
\thanks{Nagur R. C. Cherukuru is with CSIRO Environment, Acton, ACT 2601, Australia (e-mail: nagur.cherukuru@csiro.au).}%
\thanks{S. L. Kesav Unnithan is with CSIRO Space and Astronomy, Acton, ACT 2601, Australia (e-mail: kesav.unnithan@csiro.au).}%
\thanks{Tim J. Malthus, Gemma Kerrisk, and Faisal Islam are with CSIRO Environment, Dutton Park, QLD 4102, Australia (e-mail: tim.malthus@csiro.au; gemma.kerrisk@csiro.au; faisalislam0407@gmail.com).}%
\thanks{Xiubin Qi is with CSIRO Space and Astronomy, Kensington, WA 6151, Australia (e-mail: xiubin.qi@csiro.au).}%
\thanks{Tisham Dhar is with CSIRO Space and Astronomy, Adelaide, SA 5000, Australia (e-mail: tisham.dhar@csiro.au).}%
\thanks{Mark J. Doubell is with Office for Algal Bloom Research, South Australian Research and Development
Institute, Aquatic Sciences, West Beach, SA 5024, Australia (e-mail: mark.doubell@sa.gov.au).}}

\maketitle

\begin{abstract}
Hyperspectral in situ sensing has shown promise in retrieving aquatic \hl{water quality (WQ)} parameters, such as total suspended solids (TSS), dissolved organic carbon (DOC), and total chlorophyll-a (TChl-a), for cost-effective management of coastal WQ. However, generalising such retrieval algorithms across water bodies remains challenging, as the relationship between remote sensing reflectance ($R_{rs}$) and \hl{WQ} parameters can vary considerably from one region to another due to regional distinctions in environmental conditions and biogeochemistry that lead to different \hl{WQ} ranges and bio-optical properties. In this study, we propose a two-stage physics-aware meta-learning framework for retrieving coastal \hl{WQ} parameters from near-surface $R_{rs}$ observations. \hl{Unlike purely data-driven approaches trained solely on paired observations, the proposed framework incorporates physical knowledge through a two-stage learning process. In the first stage, a region-agnostic base model is pretrained using meta-learning on synthetic WQ--$R_{rs}$ samples generated by a bio-optical forward model parameterised with an in situ bio-optical spectral library. In the second stage, the pretrained model is fine-tuned using local samples to adapt to regional conditions.} To evaluate the proposed approach, we collected in situ hyperspectral $R_{rs}$ and \hl{WQ} measurements from five geographically distinct sites in Australian coastal waters. Our experimental results suggest the following: (1) the \hl{WQ} parameters and their corresponding hyperspectral $R_{rs}$ signatures exhibit clear regional distinctions among the experimental sites; (2) the synthetic dataset, generated under guidance of the bio-optical forward model and the bio-optical spectral library, is physically plausible and closely aligned with real-world samples in both parameter distributions and inter-parameter correlations; (3) when aggregated across the experimental sites, the proposed approach achieves improvements of \hl{3.3\%}, \hl{6.4\%}, and 14.1\% in \hl{site-mean} Log-$R^2$ for TSS, DOC, and TChl-a, respectively, while reducing Log-MAE by \hl{17.5\%}, 2.7\%, and 14.8\% relative to the best benchmark model; and (4) time series of in situ measured and model-predicted \hl{WQ} parameters show good agreement in both magnitude and temporal dynamics, highlighting the potential of in situ hyperspectral sensing as a cost-effective solution for continuous time-series monitoring of coastal WQ. These results demonstrate that the proposed physics-aware meta-learning framework provides a robust and adaptable approach for accurate \hl{WQ} retrieval using hyperspectral in situ sensing.
\end{abstract}


\section{Introduction}
\label{sec:introduction}

Water quality \hl{(WQ)} is an important indicator of aquatic ecosystem health \cite{zhi2024deep, Lehmann2023, unnithan2025}. Degradation of \hl{WQ} in aquatic systems undermines ecosystem functions and services \cite{medina2026data}. Timely, accurate, and continuous monitoring of \hl{WQ} is therefore critical for effective management and mitigation of environmental impacts, particularly in Australian coastal waters where many regions are vulnerable to anthropogenic pressures and climate-driven change \cite{schaffelke2012water, unnithan2025, guo2025decadal}. 

Key indicators of WQ include total suspended solids (TSS), dissolved organic carbon (DOC), total chlorophyll-a (TChl-a), and their inherent optical properties (IOPs) \cite{OShea2023,lou2025variational,LUO2025104761}. TSS represents all particulate matter suspended in the water column, including phytoplankton and non-algal particles (NAP) \cite{baker1984effect}. TChl-a is widely used as an indicator of phytoplankton biomass \cite{huot2007relationship}, while NAP comprises the non-phytoplankton fraction of TSS \cite{babin2003variations}. In addition to suspended particulates, dissolved components also play a critical role in the bio-optical properties of natural waters \cite{nelson2013global}. Dissolved organic carbon (DOC) represents the carbon content of the diverse pool of dissolved organic matter (DOM) \cite{hansell1998deep, hansell2009dissolved}, while its optically active fraction, known as chromophoric dissolved organic matter (CDOM), is widely used as a proxy for the source and compositional characteristics of DOM \cite{del2004spatial}. \hl{TSS, DOC, and TChl-a were selected in this study because they provide complementary information on suspended particulate matter, dissolved organic carbon, and phytoplankton biomass, respectively, supporting the assessment of coastal WQ.} The IOPs, including absorption and scattering coefficients, serve as the optical fingerprints of \hl{WQ} parameters, characterising the intrinsic interactions between water constituents and light \cite{Lee2002}. The mass-normalised IOPs, known as specific IOPs (SIOPs), describe how efficiently substances in water absorb or scatter light per unit concentration \cite{Lee2002}. Accurate local calibration of SIOPs is essential, as they may vary from one region to another due to differences in the composition and physiology of phytoplankton community, the size distributions of suspended particles, and the sources and processing of \hl{WQ} constituents \cite{werdell2013,cherukuru2020semi}. Collectively, these properties provide a basis for assessing \hl{WQ} by optical means.

Traditional \hl{WQ} monitoring relies on laboratory analysis of grab samples to derive \hl{WQ} parameters (\emph{e.g.}, \cite{cherukuru2017impact}). While these methods are accurate and well-established, they are often labour-intensive, costly, and logistically constrained \cite{unnithan2025,LUO2025104761}. Remote sensing has emerged as a cost-effective approach to retrieve key \hl{WQ} parameters remotely \cite{zhi2024deep}, and hyperspectral observations are particularly well-suited for capturing the spectral signatures required to assess \hl{WQ} conditions \cite{Lehmann2023, OShea2023}. Recent advances in data-driven algorithms, particularly those leveraging rigorous machine learning and deep learning frameworks, have shown promise for improving the accuracy and robustness of \hl{WQ} parameter retrieval from hyperspectral $R_{rs}$ measurements (\emph{e.g.}, \cite{OShea2023,LUO2025104761,lou2025variational}). \hl{Recent studies have advanced multi-parameter retrieval and generalisation across optically complex waters. Pahlevan et al.~\cite{pahlevan2022simultaneous} developed mixture density networks for simultaneous retrieval of chlorophyll-a, TSS, and coloured dissolved organic matter absorption from Landsat-8, Sentinel-2, and Sentinel-3 observations. Werther et al.~\cite{werther2025generalization} evaluated probabilistic neural networks for hyperspectral retrieval of absorption properties, highlighting challenges in extrapolation beyond the training distribution and in uncertainty calibration. Cao et al.~\cite{cao2026ocrt} proposed OCRT-Net, which embeds ocean-colour radiative-transfer theory within a neural-network architecture for retrieving optical water constituents. These studies provide important foundations for multi-parameter, IOP-based, and physics-informed retrieval.}

Previous studies have identified that algorithms developed and validated within a particular region for hyperspectral \hl{WQ} retrieval may exhibit limited transferability to other regions with optically distinct waters \cite{melin2015optically,spyrakos2018optical,zhou2026spectral,guo2025spatioformer}. This lack of generalisation stems from the fact that the relationship between hyperspectral $R_{rs}$ and \hl{WQ} parameters is not universal, but rather depends on regional differences in environmental conditions and biogeochemistry that lead to distinct \hl{WQ} ranges and SIOP characteristics \cite{werdell2013,neil2019global,OShea2023}. Several seminal works have observed such regional distinctions in the relationship between \hl{WQ} parameters and hyperspectral $R_{rs}$, and proposed strategies to address it. For example, Mao et al.~\cite{mao2012regional} reported that some regions in the East China Sea exhibit extremely high total suspended matter (TSM), exceeding the valid range of standard retrieval algorithms. To address this, the authors proposed a complex-proxy TSM model that blends four $R_{rs}$-based indices to transform the non-linear TSM--$R_{rs}$ relationship into a quasi-linear one, enabling robust retrievals across the full concentration range. Ogashawara et al. \cite{ogashawara2016re} observed that the Funil and Itumbiara reservoirs in Brazil are dominated by CDOM as the primary optically active constituent, whereas standard quasi-analytical algorithms (QAAs), originally designed for phytoplankton-dominated waters, failed to accurately retrieve the IOPs. To overcome this limitation, the QAA was re-parameterised by Ogashawara et al. \cite{ogashawara2016re} to suit these CDOM-dominated conditions. In a study to retrieve chlorophyll-a concentrations in 185 inland and coastal water bodies worldwide, Neil et al. \cite{neil2019global} demonstrated that improved retrieval accuracies could be achieved when algorithms are optimised for individual optical water types. These regional bio-optical distinctions contribute to the non-unique inverse problem \cite{OShea2023}, as the same (or nearly identical) spectral signatures of $R_{rs}$ may correspond to different values of \hl{WQ} parameters across water bodies. Therefore, robust algorithms for retrieving \hl{WQ} parameters require explicit accommodation of such regional variations in bio-optical conditions.

In addition to handling regional bio-optical distinctions, data-driven \hl{WQ} retrieval from hyperspectral $R_{rs}$ measurements faces further challenges. The first challenge is that data-driven algorithms often require more data for training than empirical and semi-analytical models, while the in situ sampling process is labour\hl{-intensive} and time-consuming. Several strategies have been implemented to handle this challenge, including: (1) self‑supervised pretraining on unlabelled hyperspectral $R_{rs}$ spectra is used to learn general representations, thereby reducing the number of labelled \hl{WQ}--$R_{rs}$ pairs required for subsequent model fine‑tuning \cite{LUO2025104761}; and (2) aggregating samples from multiple sources into large, curated datasets, \emph{e.g.} GLORIA \cite{Lehmann2023}, to facilitate the development of generalisable and adaptive models \cite{OShea2023}. The second challenge is that existing data-driven algorithms for hyperspectral retrieval of \hl{WQ} parameters often rely on observational data only for model training without integrating knowledge of the underlying physical principles of underwater optics. This may limit the generalisability and interpretability of the developed models.

\hl{Meta-learning aims to acquire transferable knowledge across a collection of related tasks, enabling a model to adapt rapidly to a new task using limited labelled data \cite{finn2017model}. Within remote sensing, meta-learning has primarily been investigated for few-shot image-classification problems, including hyperspectral image classification \cite{zhang2023bayesian}. Its application to ocean-colour remote sensing, however, remains largely unexplored. Existing approaches for improving the generalisability of aquatic WQ retrieval have instead predominantly relied on self-supervised pretraining \cite{LUO2025104761}, cross-regional transfer learning \cite{zhou2026spectral}, pooled multi-region training, or calibration according to optical water types \cite{neil2019global}.}

\hl{Motivated by the aforementioned methodological gaps}, we propose a two-stage training framework in this study: (1) a physics-aware pretraining stage, and (2) a region-specific adaptation stage. In the pretraining stage, a physics-based bio-optical model is used to synthesise a \hl{WQ}--$R_{rs}$ dataset from a bio-optical library of in situ measured \hl{WQ} parameters and IOPs/SIOPs with broad representativeness of Australian coastal waters. This physics-guided synthetic dataset is applied to train a region-agnostic base model within a physics-aware meta-learning framework. In the adaptation stage, the pretrained base model is subsequently adapted to individual regions using in situ measured \hl{WQ}--$R_{rs}$ samples specific to each region. The base model can regulate regional adaptation by leveraging physical knowledge learned from pretraining. The proposed algorithm is designed to incorporate the following features that distinguish it from existing data-driven approaches to \hl{WQ} retrieval from hyperspectral $R_{rs}$ observations:
\begin{enumerate}
  \item Integration of physical principles within a data-driven deep learning framework, enabling more robust modelling of the \hl{WQ}--$R_{rs}$ relationship and \hl{more physically plausible WQ retrieval};
  \item Adaptability to account for region-specific bio-optical distinctions in the \hl{WQ}--$R_{rs}$ relationship, thereby improving retrieval accuracies within each region.
\end{enumerate}

Supported by CSIRO's AquaWatch Australia initiative\footnote{\url{https://www.csiro.au/en/about/challenges-missions/AquaWatch}}, we established four experimental sites in Australian coastal waters, namely the Fitzroy Estuary and Keppel Bay sites in Queensland, the Boston Bay site in South Australia, and the Cockburn Sound site in Western Australia. Time-series measurements of in situ \hl{WQ} parameters and hyperspectral $R_{rs}$ were collected at these sites across multiple seasons from 2022 to 2025. In situ measurements were also sourced from a long-term monitoring site, namely the Lucinda Jetty Coastal Observatory in Queensland, from 2019 to 2023. These five sites cover several climate zones, including tropical, subtropical, and temperate, and represent a range of hydrodynamic regions, including riverine, estuarine, and coastal. As a result, the waters at these sites exhibit distinct bio-optical characteristics. 

The proposed model for retrieving \hl{WQ} parameters was evaluated using data collected from the aforementioned sites. Through quantitative analysis of the evaluation results, we aim to address primarily the following questions:
\begin{enumerate}

    \item Is there a clear regional distinction among the experimental sites in terms of \hl{WQ} parameters and their hyperspectral $R_{rs}$ signatures?

    \item How does the proposed approach perform in hyperspectral retrieval of \hl{WQ} parameters compared to benchmark algorithms?
    
\end{enumerate}

The rest of this article is organised as follows. Section~\ref{sec:study_sites_and_datasets} describes the study sites, in situ hyperspectral $R_{rs}$ and \hl{WQ} measurements, and the bio-optical spectral library used in this study. Section~\ref{sec:methods} details the proposed physics-aware meta-learning framework, including the bio-optical forward model, synthetic data generation, model pretraining, and region-specific adaptation. Section~\ref{sec:results_and_discussions} presents the experimental results and discusses the retrieval performance, regional variability, and time-series monitoring capabilities of the proposed approach. The key findings, limitations, and future directions are also discussed in Section~\ref{sec:results_and_discussions}. Finally, Section~\ref{sec:conclusion} concludes this article with a summary overview of this study. To facilitate readability, Table~\ref{tab:symbols} lists the commonly used symbols and acronyms in this article.

\begin{table}[!t]
\centering
\caption{Symbols and acronyms used in this study.}
\label{tab:symbols}

\begin{tabularx}{\columnwidth}{
>{\raggedright\arraybackslash}p{1.1cm}
>{\raggedright\arraybackslash}p{1.2cm}
X}
\toprule
Symbol & Unit & Definition \\
\midrule

\multicolumn{3}{l}{\textit{\hl{Water Quality (WQ)} Parameters}}\\
TSS & mg/L & Total suspended solids concentration\\
TChl-a & $\mu$g/L & Total chlorophyll-a concentration\\
NAP & mg/L & Non-algal particles concentration\\
DOC & mg/L & Dissolved organic carbon concentration\\

\addlinespace
\multicolumn{3}{l}{\textit{Inherent Optical Properties (IOPs)}}\\
$a$ & m$^{-1}$ & Total absorption coefficient\\
$a_w$ & m$^{-1}$ & Absorption coefficient of pure water\\
$a_d$ & m$^{-1}$ & Absorption coefficient of NAP\\
$a_{ph}$ & m$^{-1}$ & Absorption coefficient of phytoplankton\\
$a_y$ & m$^{-1}$ & Absorption coefficient of Chromophoric dissolved organic matter (CDOM)\\
$S_y$ & nm$^{-1}$ & Spectral slope of CDOM absorption coefficient\\
$b_b$ & m$^{-1}$ & Total backscattering coefficient\\
$b_{bw}$ & m$^{-1}$ & Backscattering coefficient of pure water\\
$b_{bp}$ & m$^{-1}$ & Backscattering coefficient of particulate matter\\
$S_{bbp}$ & -- & Spectral slope of particulate backscattering coefficient\\

\addlinespace
\multicolumn{3}{l}{\textit{Specific IOPs (SIOPs)}}\\
$a_d^{*}$ & m$^{2}$ g$^{-1}$ & TSS-normalised NAP absorption coefficient\\
$a_{ph}^{*}$ & m$^{2}$ mg$^{-1}$ & TChl-a-normalised phytoplankton absorption coefficient\\
$a_y^{*}$ & m$^{2}$ g$^{-1}$ & DOC-normalised CDOM absorption coefficient\\
$b_{bp}^{*}$ & m$^{2}$ g$^{-1}$ & TSS-normalised particulate backscattering coefficient\\

\addlinespace
\multicolumn{3}{l}{\textit{Apparent Optical Properties (AOPs)}}\\
$R_{rs}$ & sr$^{-1}$ & Above-water remote sensing reflectance\\
$r_{rs}$ & sr$^{-1}$ & Subsurface remote sensing reflectance\\

\addlinespace
\multicolumn{3}{l}{\textit{Ancillary Variables}}\\
$T$ & $^\circ$C & Water temperature\\
$S$ & -- & Water salinity\\

\bottomrule
\end{tabularx}
\end{table}
\section{Study sites and datasets}
\label{sec:study_sites_and_datasets}

\subsection{Study sites}

\begin{figure*}[!t]
\centering
\includegraphics[width=0.8\textwidth]{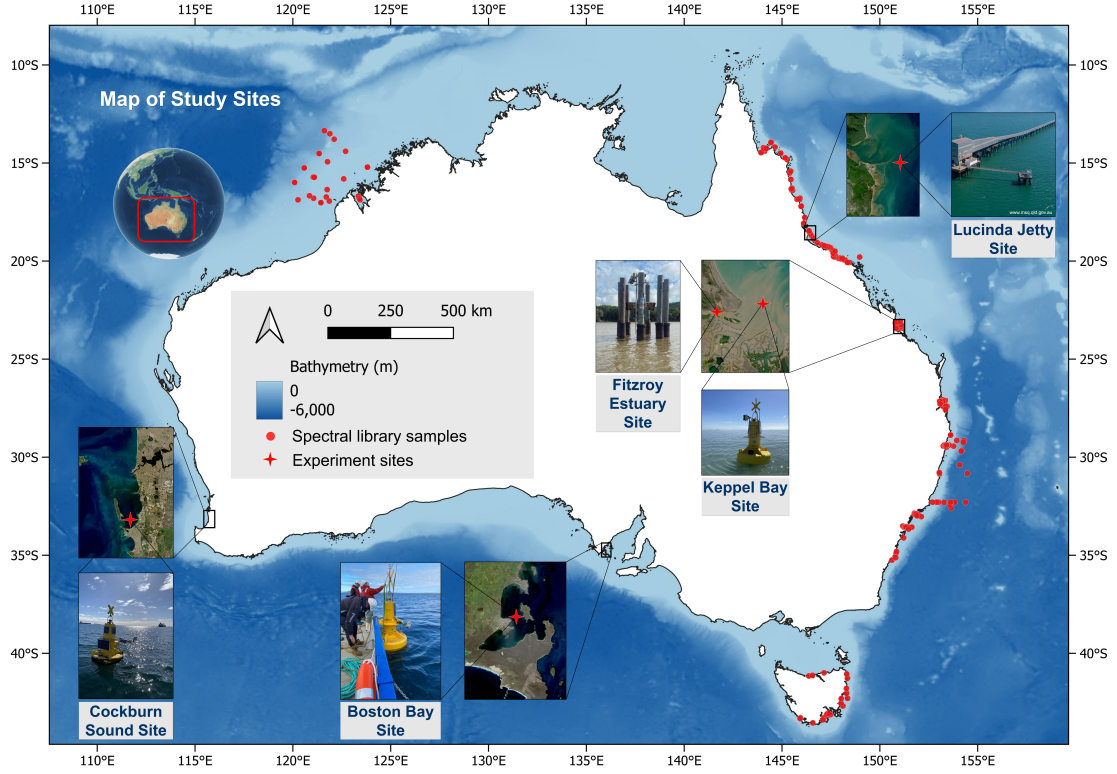}
\caption{This study was conducted across five experimental sites in Australian coastal waters: Fitzroy Estuary, Keppel Bay, Boston Bay, Cockburn Sound, 
and Lucinda Jetty. The map also shows the locations (red dot points) where the bio-optical spectral library samples were collected. Insets show satellite images of the experimental sites and adjacent areas, along with photos of the instruments taken during or after deployment. \label{fig:location}}
\end{figure*}

This study focused on five sites located in Australian coastal waters, as shown in Fig.~\ref{fig:location}. The Fitzroy Estuary and Keppel Bay sites are located near the mouth of Fitzroy River, Queensland, with the former situated inside the estuary and the latter downstream in open coastal waters in Keppel Bay. The Boston Bay site is located in the strait between Boston Island and the mainland town of Port Lincoln, within Boston Bay in Spencer Gulf, South Australia. The Cockburn Sound site is located within Cockburn Sound, Western Australia, between Garden Island and the Perth metropolitan coast. 
The last site, Lucinda Jetty, is located in the coastal waters of the Great Barrier Reef World Heritage Area, close to the Herbert River Estuary and the Hinchinbrook Channel in Queensland. These geographically distinct sites span tropical, subtropical, and temperate climate zones, and are influenced to varying degrees by terrestrial freshwater discharge as reflected by their salinity levels (Table~\ref{tab:sites}).

\begin{table}[!t]
\centering
\caption{List of experimental sites.}
\label{tab:sites}

\begin{tabularx}{\columnwidth}{
>{\raggedright\arraybackslash}p{2.0cm}
>{\raggedright\arraybackslash}X
>{\raggedright\arraybackslash}p{1.6cm}
>{\centering\arraybackslash}X
>{\centering\arraybackslash}X
}
\toprule
Site name & State & Nominal geo-coordinates & Climate zone & Mean salinity \\
\midrule

Fitzroy Estuary
& Queensland
& 23\textdegree{}30$'$07$''$S, 150\textdegree{}48$'$03$''$E
& Subtropical
& 27.1 \\

Keppel Bay
& Queensland
& 23\textdegree{}27$'$54$''$S, 150\textdegree{}56$'$39$''$E
& Subtropical
& 36.1 \\

Boston Bay
& South Australia
& 34\textdegree{}43$'$03$''$S, 135\textdegree{}54$'$02$''$E
& Temperate
& 36.3 \\

Cockburn Sound
& Western Australia
& 32\textdegree{}15$'$03$''$S, 115\textdegree{}43$'$42$''$E
& Temperate
& 35.5 \\

Lucinda Jetty
& Queensland
& 18\textdegree{}31$'$11$''$S, 146\textdegree{}23$'$10$''$E
& Tropical
& 34.8 \\

\bottomrule
\end{tabularx}
\end{table}

\subsection{Near-surface hyperspectral $R_{rs}$ observations}
\label{ssec:hyperspectral_reflectance}

\begin{figure*}[!t]
\centering
\includegraphics[width=0.8\textwidth]{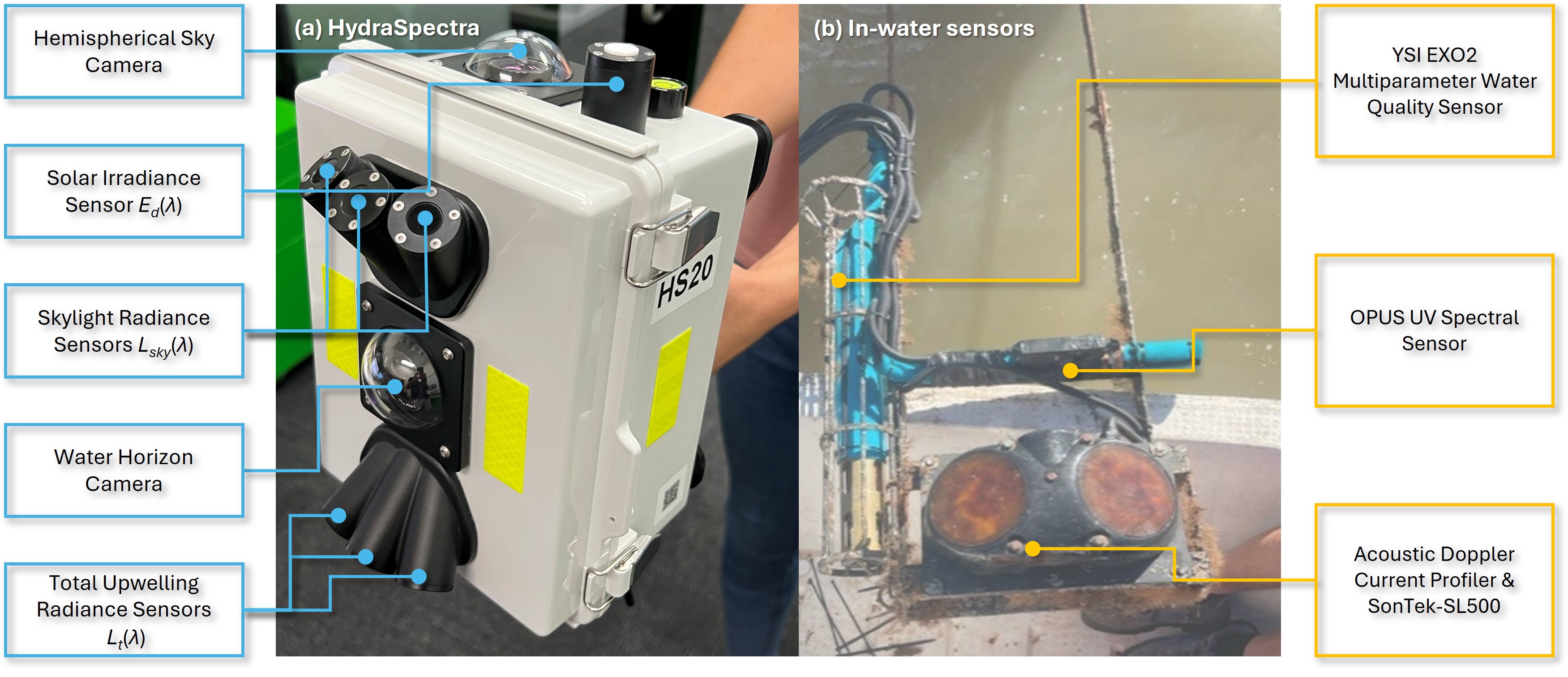}
\caption{In situ sensors deployed at the experimental sites of Fitzroy Estuary, Keppel Bay, Boston Bay, and Cockburn Sound. (a) HydraSpectra instrument for recording water-leaving hyperspectral $R_{rs}$. (b) In-water sensors for collecting \hl{WQ} measurements and monitoring water conditions. \label{fig:sensors}}
\end{figure*}

A HydraSpectra instrument\footnote{\url{https://www.csiro.au/en/research/technology-space/data/hydraspectra}}
(Fig.~\ref{fig:sensors}a) was deployed to measure hyperspectral water-leaving $R_{rs}$ at each of the Fitzroy Estuary, Keppel Bay, Boston Bay, and Cockburn Sound sites. These instruments were calibrated in the laboratory prior to deployment. \hl{They} recorded time-series hyperspectral measurements at 10-minute intervals, including downwelling solar irradiance ($E_d(\lambda)$), \hl{total upwelling radiance ($L_t(\lambda)$)}, and diffuse skylight radiance ($L_{sky}(\lambda)$). These measurements were used to calculate the water-leaving $R_{rs}$ following the method described in \hl{Mobley}~\cite{mobley1999estimation}. \hl{A subsequent residual surface-reflection correction was applied using the similarity-spectrum method of Ruddick et al.~\cite{Ruddick2006}.} To ensure a high signal-to-noise ratio for the HydraSpectra sensor, we selected $R_{rs}$ observations acquired around local noontime (10:00am -- 2:00pm), when solar irradiance is usually stronger than at other times of the day, for our analysis. \hl{Built into the HydraSpectra instrument, two RGB cameras operated concurrently with the spectral sensors to capture true-colour images of the hemispherical sky and the water horizon. These images were reviewed by a technician during the Quality Assurance and Quality Control (QA/QC) process to identify and exclude low-quality $R_{rs}$ observations, such as those affected by sun glint, bad weather, or wildlife interference. Only high-fidelity spectra were retained for analysis. For the retrieval analysis, the corrected HydraSpectra $R_{rs}$ spectra were restricted to 400--700 nm, with a spectral resolution of approximately 3 nm and a sampling interval of 1 nm. Wavelengths outside this range were excluded from the retrieval analysis because of low signal-to-noise ratios.} These HydraSpectra instruments were regularly serviced and calibrated by the CSIRO servicing team. At the Lucinda Jetty site, water-leaving hyperspectral $R_{rs}$ was derived from measurements acquired using a Hyperspectral Ocean Colour Radiometer (HyperOCR). These data are available through the Integrated Marine Observing System (IMOS) via the Australian Ocean Data Network (AODN) portal\footnote{\url{https://thredds.aodn.org.au/thredds/catalog/IMOS/SRS/OC/LJCO/HyperOCR-hourly/catalog.html}}. Similar to the HydraSpectra data, the spectral range from 400 to 700~nm in the HyperOCR measurements was used for analysis.

\subsection{In situ \hl{WQ} parameter measurements}
\label{ssec:water_quality_measurements}

At the experimental sites of Fitzroy Estuary, Keppel Bay, Boston Bay, and Cockburn Sound, a YSI EXO2 multiparameter sonde and an OPUS UV Spectral Sensor (Fig.~\ref{fig:sensors}b) were installed at a nominal depth of 1~m below the water surface, at the same location as the above-water HydraSpectra deployment. Both sensors were configured to record \hl{WQ} measurements at 10 min intervals. \hl{To ensure data quality, these sensors underwent regular servicing during the study period. In cases of malfunction, emergency repairs were performed as soon as possible to minimise data interruptions. A total of 56 grab samples of water from these in situ locations were collected during the study period and transported to the CSIRO Hydrochemistry Laboratories in Hobart, Tasmania, for analysis. These grab samples were used to calibrate and verify the EXO2 and OPUS measurements. Relative to the independent laboratory reference measurements, the calibrated sensors yielded root-mean-square errors of 0.74 mg/L for TSS, 0.55 mg/L for DOC, and 0.61 $\mu$g/L for TChl-a.} For the Lucinda Jetty site, grab samples of water were collected at the same location as the above-water HyperOCR deployment, and are available through the IMOS AODN portal\footnote{\url{https://thredds.aodn.org.au/thredds/catalog/IMOS/SRS/OC/BODBAW/catalog.html}}. \hl{The WQ parameters at Lucinda Jetty were measured via laboratory analysis of grab samples.} The number of in situ \hl{WQ} samples and the corresponding sampling period at each experimental site are listed in Table \ref{tab:datasets}.

\begin{table}[!t]
\centering
\caption{Summary of in situ \hl{WQ} samples collected at each experimental site.}
\label{tab:datasets}

\begin{tabularx}{\columnwidth}{
    >{\raggedright\arraybackslash}p{2cm}
    >{\centering\arraybackslash}p{2cm}
    >{\raggedright\arraybackslash}X
}
\toprule
Site & No. of Samples & Sampling Period \\
\midrule
Fitzroy Estuary & 296 & 26 Apr 2023 -- 23 Sep 2024 \\
Keppel Bay & 377 & 1 Jun 2023 -- 28 Nov 2024 \\
Boston Bay & 774 & 1 Sep 2022 -- 2 Feb 2025 \\
Cockburn Sound & 172 & 15 Jul 2023 -- 18 Feb 2024 \\
Lucinda Jetty & 104 & 12 Nov 2019 -- 1 Nov 2023 \\
\bottomrule
\end{tabularx}
\end{table}

\subsection{Bio-optical spectral library}
\label{ssec:bio_optical_spectral_library}

We compiled a bio-optical spectral library using 247 in situ bio-optical samples collected from several regions along Australia's coastline, including coastal waters in Queensland, New South Wales, Tasmania, and Western Australia, as shown in Fig.~\ref{fig:location}. These in situ bio-optical measurements cover a wide range of optical water types and hydro-optical conditions with broad representativeness of Australian coastal waters. The library is built upon multiple field campaigns conducted between 2002 and 2022, including: (1) seven voyages in the inshore estuarine, lagoonal, and reef waters of the Great Barrier Reef during four dry and one wet tropical seasons between October 2002 and September 2005 \cite{oubelkheir2006using,blondeau2009bio}; (2) a voyage in Tasmanian coastal waters from 21 to 30 May 2007 \cite{cherukuru2014influence}; (3) a voyage in the western Tasman Sea, off the New South Wales coast, conducted from 15 to 30 October 2010 \cite{cherukuru2016physical}; (4) a voyage across the Kimberley shelf off northwest Western Australia from 14 April to 5 May 2010 \cite{cherukuru2019bio}; (5) five voyages in Moreton Bay, Queensland, in 2011 following a major flood event \cite{oubelkheir2014impact}; (6) a voyage in Princess Charlotte Bay, Queensland, conducted from 30 January to 1 February 2013 after the passage of Tropical Cyclone Oswald \cite{oubelkheir2023impact}; and (7) two voyages across three estuaries along the New South Wales coast in March and July 2022, respectively \cite{unnithan2025}. \hl{The TSS, DOC, and TChl-a concentrations for the spectral library samples were determined from laboratory analyses of grab samples following standard analytical procedures at the CSIRO Hydrochemistry Laboratories in Hobart, Tasmania.}

For each water sample in the bio-optical library, the following \hl{WQ} parameters, IOPs, and ancillary physical variables were measured: (1) water temperature (in unit of \textdegree{}C), (2) water salinity (unitless), (3) TSS (in unit of mg/L), (4) TChl-a (in unit of $\mu$g/L), (5) DOC (in unit of mg/L), (6) absorption coefficient of CDOM at 440 nm ($a_y(\lambda_{440})$, in unit of m\textsuperscript{\textminus 1}) and its spectral slope ($S_y(\lambda_{440})$, in unit of nm\textsuperscript{\textminus 1}), (7) backscattering coefficient of particulate matter at 550 nm ($b_{bp}(\lambda_{550})$, in unit of m\textsuperscript{\textminus 1}) and its spectral slope ($S_{bbp}(\lambda_{550})$, unitless), (8) absorption coefficient of particulate matter within 400--700 nm ($a_{p}(\lambda)$, in unit of m\textsuperscript{\textminus 1}), (9) absorption coefficient of NAP within 400--700 nm ($a_{d}(\lambda)$, in unit of m\textsuperscript{\textminus 1}), and (10) absorption coefficient of phytoplankton within 400--700 nm ($a_{ph}(\lambda)$, in unit of m\textsuperscript{\textminus 1}). The SIOPs were then derived, including: (1) \hl{DOC-normalised} absorption coefficient of CDOM ($a^*_y(\lambda)$, in unit of m\textsuperscript{2}~g\textsuperscript{\textminus 1}), (2) mass-specific backscattering coefficient of particulate matter ($b^*_{bp}(\lambda)$, in unit of m\textsuperscript{2}~g\textsuperscript{\textminus 1}), (3) mass-specific absorption coefficient of NAP ($a^*_{d}(\lambda)$, in unit of m\textsuperscript{2}~g\textsuperscript{\textminus 1}), and (4) mass-specific absorption coefficient of phytoplankton ($a^*_{ph}(\lambda)$, in unit of m\textsuperscript{2}~mg\textsuperscript{\textminus 1}). The hyperspectral water-leaving reflectance ($R_{rs}(\lambda)$, in unit of sr\textsuperscript{\textminus 1}) for each water sample was calculated using the bio-optical forward model described in Subsection~\ref{ssec:bio_optical_model}.

\section{Methods}
\label{sec:methods}

\subsection{Overview}

\begin{figure*}[!t]
\centering
\includegraphics[width=0.8\textwidth]{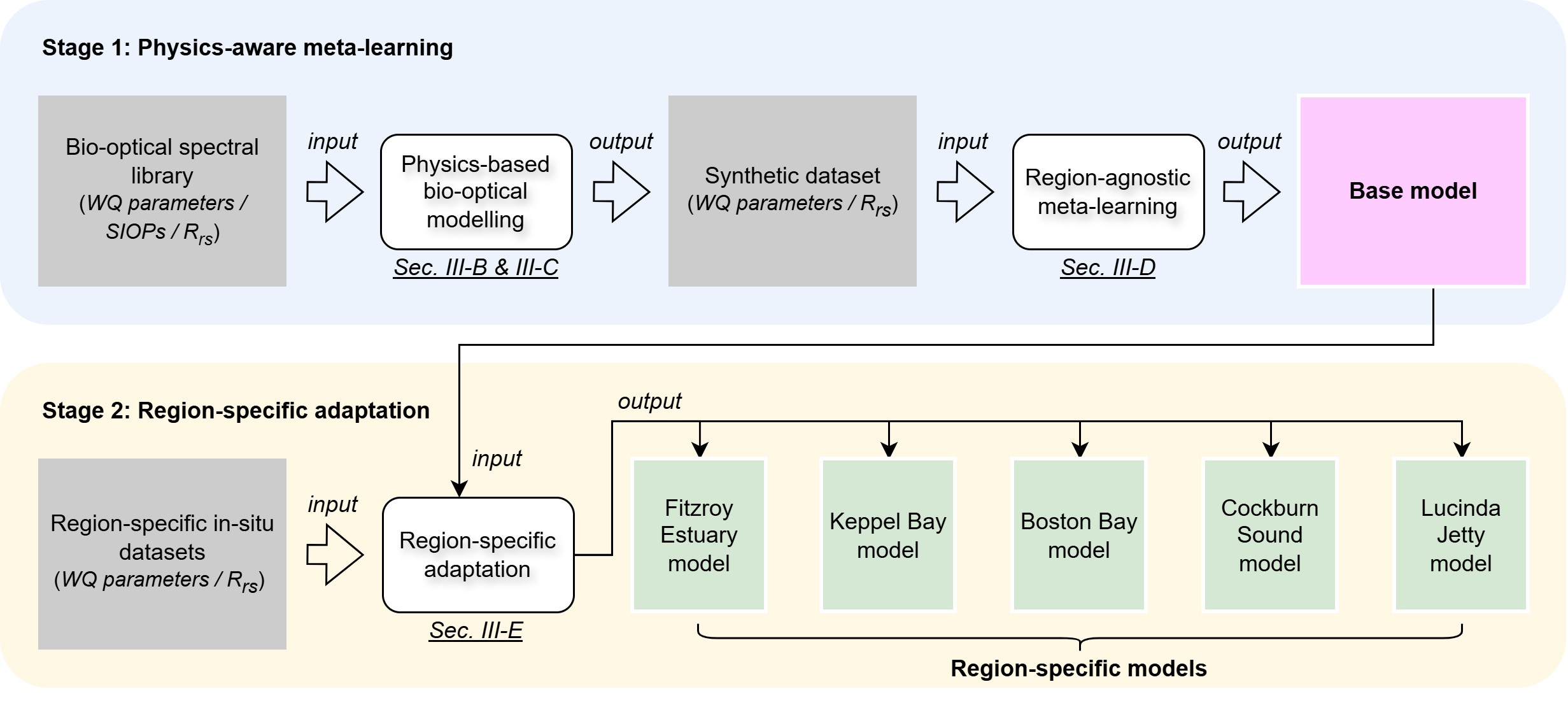}
\caption{Flowchart of the proposed physics-aware meta-learning approach for hyperspectral retrieval of \hl{WQ} parameters. It consists of two major stages: (1) a physics-aware pretraining stage (detailed in Subsections \ref{ssec:bio_optical_model}, \ref{ssec:synthetic_data_generation}, and \ref{ssec:pretraining}), and (2) a region-specific adaptation stage (detailed in Subsection \ref{ssec:adaptation}). Please refer to the nomenclature in Table \ref{tab:symbols} for definitions of symbols and acronyms. \label{fig:flowchart}}
\end{figure*}

We aim to develop a physics-aware meta-learning approach for retrieving \hl{WQ} parameters from in situ hyperspectral $R_{rs}$ measurements. Given that the relationship between $R_{rs}$ and \hl{WQ} parameters may vary from one region to another due to distinct regional bio-optical characteristics, we propose a two-stage framework for model training, as shown in Fig. \ref{fig:flowchart}. In the first stage, a physics-based bio-optical forward model was adopted to simulate a large physics-guided synthetic dataset linking \hl{WQ} parameters and their SIOPs with hyperspectral $R_{rs}$. The simulation was conducted based on the bio-optical spectral library. Then, a region-agnostic base model was trained with the synthetic dataset using meta-learning. In the second stage, the base model was fine-tuned for each target region using in situ \hl{WQ}--$R_{rs}$ samples, resulting in a region-specific model for each region. This two-stage framework was designed to integrate physical modelling with data-driven learning, with the potential to accommodate regional bio-optical distinctions. The proposed framework is described in detail in the subsequent Subsections.

\subsection{Bio-optical forward model}
\label{ssec:bio_optical_model}

We use a physics-based bio-optical model to link \hl{WQ} parameters and their  SIOPs with hyperspectral $R_{rs}$. Specifically, the total spectral absorption coefficient of the water, $a$, is expressed as:
\begin{equation}
a(\lambda,T,S)
=
a_w(\lambda,T,S)
+a_d(\lambda)
+a_y(\lambda)
+a_{ph}(\lambda),
\end{equation}
where $a_w$, $a_d$, $a_y$, and $a_{ph}$ represent absorption by water, NAP, CDOM, and phytoplankton, respectively.

To parameterise $a_d$, $a_y$, and $a_{ph}$ using \hl{WQ} measurements, we derived empirical SIOPs by normalising these component absorption coefficients by the corresponding TSS, DOC, and TChl-a concentrations:
\begin{equation}
\label{eq:abs_coef}
\begin{split}
a(\lambda,T,S) = &a_w(\lambda,T,S)
+\mathrm{TSS} \times a_d^*(\lambda)\\
&+\mathrm{DOC} \times a_y^*(\lambda)
+\mathrm{TChl\mbox{-}a} \times a_{ph}^*(\lambda).
\end{split}
\end{equation}
Here, $a_d^*$, $a_y^*$, and $a_{ph}^*$ are the TSS-normalised $a_d$, DOC-normalised $a_y$, and TChl-a-normalised $a_{ph}$, respectively. \hl{It is worth noting that $a_y^*(\lambda)=a_y(\lambda)/\mathrm{DOC}$
is an empirical DOC-normalised CDOM absorption coefficient, rather than a universally invariant SIOP.}

Given that most of our experimental sites are located in coastal waters influenced by freshwater--seawater mixing, the spectral curve of $a_w$ is computed accounting for variations in water salinity ($S$) and temperature ($T$) to enhance accuracy:
\begin{equation}
a_w(\lambda, T, S) \;=\; a_w^{\text{ref}}(\lambda) 
+ \bigl(T - T_w^{\text{ref}}\bigr)\times\Psi_T(\lambda) 
+ S \times \Psi_S(\lambda),
\end{equation}
where $a_w^{\text{ref}}$ is the reference water absorption spectrum, adapted from \hl{Pope and Fry}~\cite{pope1997absorption}; $T_w^{\text{ref}}=22^{\circ}$C is the reference temperature; $\Psi_T$ and $\Psi_S$ are the correction coefficients for temperature and salinity, respectively, adapted from \hl{Röttgers et al.}~\cite{rottgers2014temperature}.

The total spectral backscattering coefficient, $b_b$, is expressed as:
\begin{equation}
b_b(\lambda,S)
=
b_{bw}(\lambda,S)
+
\mathrm{TSS} \times b_{bp}^*(\lambda),
\end{equation}
where $b_{bp}^*$ is the TSS-specific particulate backscattering coefficient. Following the molecular scattering formulation of \hl{Boss and Pegau}~\cite{boss2001relationship}, the water volume-scattering function is calculated as:
\begin{equation}
A(\lambda,S)
=
1.38 \times 10^{-4}
\times
\left(\frac{\lambda}{500}\right)^{-4.32}
\times
\left(1+0.3 \times \frac{S}{37}\right).
\end{equation}
The backscattering coefficient of water, $b_{bw}$, is then
obtained by integrating the volume-scattering function over the backward hemisphere:
\begin{equation}
b_{bw}(\lambda,S)
=
2\pi \times A(\lambda,S)
\times
\left[
1+\frac{1-\delta}{3(1+\delta)}
\right],
\end{equation}
where $\delta=0.09$ is the molecular depolarisation ratio.

Following a standard radiative-transfer approximation \cite{Lee2002}, the subsurface $R_{rs}$ at $0^-$,  \(r_{rs}\), is related to the total backscattering albedo, \(u\), by:
\begin{equation}
r_{rs}(\lambda)=g_0\times u(\lambda)+g_1\times u^2(\lambda),
\label{eq:rrs_ut}
\end{equation}
where \(g_0=0.082\) and \(g_1=0.17\) are empirical coefficients derived from the HydroLight radiative transfer model \cite{mobley1989numerical}, and \(u\) is expressed as:
\begin{equation}
u(\lambda)=\frac{b_b(\lambda)}{a(\lambda)+b_b(\lambda)}.
\label{eq:ut}
\end{equation}

To obtain the above-water $R_{rs}$ at $0^+$, $R_{rs}$, we apply a standard air–water interface correction:
\begin{equation}
R_{rs}(\lambda)=\frac{0.52\times r_{rs}(\lambda)}{1-1.7\times r_{rs}(\lambda)}.
\label{eq:Rrs_rrs}
\end{equation}
where the correction coefficients \(0.52\) and \(1.7\) follow the values given in \hl{Lee et al.}~\cite{Lee2002}.

\subsection{Synthetic data generation}
\label{ssec:synthetic_data_generation}

We generate a physics-based synthetic dataset with bio-optical modelling. 
For a given set of input parameters (TSS, DOC, TChl-a, $a^*_d$, $a^*_y$, $a^*_{ph}$, $b^*_{bp}$, $T$, and $S$), the bio-optical forward model, as described in Subsection \ref{ssec:bio_optical_model}, is able to simulate the corresponding hyperspectral $R_{rs}$. Randomly selected sets of input values may result in physically unrealistic simulations. To enable physically consistent and plausible simulations, the values for simulation inputs are sampled with the guidance of the bio-optical spectral library. Specifically, we first characterise the statistical distribution of the spectral library measurements and then draw simulation inputs from this distribution, aiming to align the sampled values with the observed measurements. These sampled simulation inputs are subsequently passed to the bio-optical forward model to generate hyperspectral $R_{rs}$. This procedure is detailed in the following paragraphs.

We first apply two preprocessing steps to the spectral library measurements before modelling their statistical distribution. The first step is to transform the values of TSS, DOC, TChl-a, $a^*_d$, $a^*_y$, $a^*_{ph}$, and $b^*_{bp}$ into the $\log_{10}$ scale. Because the spectral library measurements of these \hl{WQ} parameters and SIOPs are skewed towards small values, $\log_{10}$-scaling helps normalise their distributions and stabilise subsequent modelling, as recommended by \hl{Seegers et al.}~\cite{seegers2018performance}. We denote the $\log_{10}$-scaled TSS, DOC, and TChl-a as $\mathbf{x}_{\text{TSS}}, \mathbf{x}_{\text{DOC}}, \mathbf{x}_{\text{TChl-a}} \in \mathbb{R}^N$, where $N=247$ is the number of spectral library measurements. Similarly, the $\log{10}$-scaled $a^*_d$, $a^*_y$, $a^*_{ph}$, and $b^*_{bp}$ are denoted as $\mathbf{A}_d, \mathbf{A}_y, \mathbf{A}_{ph}, \mathbf{B}_{bp} \in \mathbb{R}^{M\times N}$, with $M=301$ spectral bands covering 400--700 nm with a 1 nm interval. The $T$ and $S$ measurements are not strongly skewed and therefore left unscaled, and are denoted as $\mathbf{x}_{T}, \mathbf{x}_{S}\in \mathbb{R}^N$.

The second preprocessing step is to handle the spectral dependency of $a^*_d$, $a^*_y$, $a^*_{ph}$, and $b^*_{bp}$. Unlike other input variables to the bio-optical forward model, these SIOPs are wavelength-dependent. For each of these variables, the values across adjacent wavelengths are highly correlated, forming smooth and continuous spectral shapes in nature. To preserve these correlations across wavelengths, we transform $a^*_d$, $a^*_y$, $a^*_{ph}$, and $b^*_{bp}$ from the original spectral space into a principal component (PC) space using principal component analysis (PCA). \hl{We use the first five PCs, which together explain 98.7\%, 98.5\%, 98.3\%, and 99.0\% of the variance for $a^*_d$, $a^*_y$, $a^*_{ph}$, and $b^*_{bp}$, respectively, as a balance between information preservation and noise reduction.}
The transformed data are represented in the PC space as $\mathbf{X}_d$, $\mathbf{X}_{y}$, $\mathbf{X}_{ph}$, and $\mathbf{X}_{bbp}$. This PCA transformation captures the dominant modes of spectral variability while preserving the correlations across wavelengths, as the spectral shapes are represented by a compact set of uncorrelated PCs. 

We then model the joint statistical distribution of TSS, DOC, TChl-a, $a^*_d$, $a^*_y$, $a^*_{ph}$, $b^*_{bp}$, $T$, and $S$ for the spectral library samples based on their preprocessed values. 
We adopt an infinite Gaussian mixture model called the Dirichlet Process Bayesian Gaussian Mixture Model (DP-GMM), which is a non-parametric Bayesian framework suitable for modelling non-Gaussian multivariate distributions. Specifically, we first collate the preprocessed input variables as $\mathbf{Y} \in \mathbb{R}^{(5+4\times P) \times N}$:
\begin{equation}
    \mathbf{Y} = 
    \begin{bmatrix}
        \mathbf{x}_{\text{TSS}},
        \mathbf{x}_{\text{DOC}},
        \mathbf{x}_{\text{TChl-a}},
        \mathbf{X}_{d},
        \mathbf{X}_{y},
        \mathbf{X}_{ph},
        \mathbf{X}_{bbp},
        \mathbf{x}_{T},
        \mathbf{x}_{S}
    \end{bmatrix}^\top.
\end{equation}
Let $\mathbf{y}_n \in \mathbb{R}^{(5+4\times P)}$ denote the $n$-th measurement in $\mathbf{Y}$, and let $\mathbf{y}\in \mathbb{R}^{(5+4\times P)}$ denote a generic measurement. We then use DP-GMM to model the distribution of $\mathbf{y}$, $p(\mathbf{y})$, as an unbounded mixture of Gaussian components:
\begin{equation}
\label{eq:joint}
    p(\mathbf{y}) = \sum_{i=1}^{\infty} \pi_i \, \mathcal{N}(\mathbf{y} \mid \boldsymbol{\mu}_i, \boldsymbol{\Sigma}_i).
\end{equation}
where $\mathcal{N}(\cdot)$ 
denotes the multivariate Gaussian distribution; $\pi_i$, $\boldsymbol{\mu}_i$, and $\boldsymbol{\Sigma}_i$ are learnable parameters denoting the mixture weight, mean vector, and covariance matrix of the $i$-th Gaussian component, respectively. \hl{In this study, the DP-GMM was fitted using a truncated Dirichlet-process prior with a maximum of 10 Gaussian components, full covariance matrices, a concentration parameter of 10, and 50 k-means initialisations.}

The joint distribution $p(\mathbf{y})$ given in Eq. (\ref{eq:joint}) is derived from a large number of in situ measurements in the spectral library, and thus serves as a realistic representation of the distribution and inter-correlations of \hl{WQ} parameters and their SIOPs. We draw a total of $K$ samples from $p(\mathbf{y})$ to obtain physically plausible inputs for bio-optical modelling:
\begin{equation}
    \{\mathbf{y}^{(k)}\}_{k=1}^K \sim p(\mathbf{y}),
\end{equation}
where $\mathbf{y}^{(k)}$ is the $k$-th sample drawn from the distribution, consisting of a set of parameters, including the $\log_{10}$-scaled TSS, DOC, and TChl-a, the $\log_{10}$-scaled and PCA-transformed $a^*_d$, $a^*_y$, $a^*_{ph}$, $b^*_{bp}$, and $T$ and $S$. These parameters are then scaled and transformed back to their original scales and spaces, and denoted as $\text{TSS}^{(k)}$, $\text{DOC}^{(k)}$, $\text{TChl-a}^{(k)}$, $a^{*(k)}_d$, $a^{*(k)}_y$, $a^{*(k)}_{ph}$, $b^{*(k)}_{bp}$, $T^{(k)}$, and $S^{(k)}$. For each drawn sample, we input them into the bio-optical forward model and calculate the corresponding $R_{rs}$:
\begin{equation}
\begin{split}
R_{rs}^{(k)} = f(&\text{TSS}^{(k)}, \text{DOC}^{(k)}, \text{TChl-a}^{(k)}, \\
&a^{*(k)}_d, a^{*(k)}_y, a^{*(k)}_{ph}, b^{*(k)}_{bp},
T^{(k)}, S^{(k)}),
\end{split}
\end{equation}
where $f(\cdot)$ is the bio-optical forward model as described in Subsection \ref{ssec:bio_optical_model}. To adequately cover the distribution of the bio-optical variables, we sampled $K=10,000$ instances. Based on empirical test runs, this sample size was found to provide sufficient coverage of the parameter space, with larger sample sizes yielding only marginal gains for pretraining the base model. For each sample, the corresponding $R_{rs}$ was calculated, resulting in a synthetic dataset $\mathcal{D}$ that links \hl{WQ} parameters with $R_{rs}$:
\begin{equation}
\begin{split}
\mathcal{D} = \Bigl\{
\Bigl(
&\text{TSS}^{(k)},
\text{DOC}^{(k)},
\text{TChl-a}^{(k)},
a^{*(k)}_{d},
a^{*(k)}_{y}, \\
&a^{*(k)}_{ph},
b^{*(k)}_{bp},
T^{(k)},
S^{(k)},
R_{rs}^{(k)}
\Bigr)
\Bigr\}_{k=1}^{K}.
\end{split}
\end{equation}

\subsection{Pretraining of the base model}
\label{ssec:pretraining}

We pretrain a region-agnostic base model with the synthetic dataset $\mathcal{D}$ generated in Subsection \ref{ssec:synthetic_data_generation}. We formulate the pretraining of the base model as a meta‑learning problem and develop an approach based on the Model-Agnostic Meta-Learning (MAML) architecture \cite{finn2017model}. The base model is pretrained on a collection of tasks, where each task is formed from a subset of synthetic samples in $\mathcal{D}$. As $\mathcal{D}$ is constructed with physics-based bio-optical modelling (Subsection \ref{ssec:synthetic_data_generation}), the base model is able to learn the general physical relationship linking \hl{WQ} parameters and hyperspectral $R_{rs}$ through this pretraining process. Moreover, tasks are constructed with diverse SIOP settings, enabling the base model to adapt effectively to new regions characterised by distinct SIOP conditions. The pretraining of the base model is detailed in the following.

We denote the base model as $g_{\boldsymbol{\theta}}(\cdot):\mathbb{R}^{M}\!\to\!\mathbb{R}^{3}$ with model parameters $\boldsymbol{\theta}$, mapping a hyperspectral $R_{rs}$ spectrum $R_{rs}\in\mathbb{R}^{M}$ to the \hl{WQ} parameters TSS, DOC, and TChl-a. In our study, the number of spectral bands is $M=301$, covering 400--700 nm with a 1 nm interval. \hl{The base model was implemented as a fully connected feed-forward neural network with 301 input neurons, one for each wavelength in the input $R_{rs}$ spectrum. The input spectrum was processed through three hidden layers, with each containing 1024 neurons with ReLU activation, to predict TSS, DOC, and TChl-a through the output layer of three neurons.}

We pretrain $g(\cdot)$ on a distribution of tasks $p(\mathcal{T})$ sampled from $\mathcal{D}$. Each task \(\tau_i\) is constructed by sampling a subset \(\mathcal{D}_{i}\!\subset\!\mathcal{D}\) of \(2K_i\) samples and splitting it evenly into a support (training) set \(\mathcal{S}_i\) and a query (test) set \(\mathcal{Q}_i\), with each having \(K_i\) samples. 
The experimental sites considered in this study may present distinct SIOPs, due to differences in regional biogeochemistry, such as phytoplankton community composition and physiology, suspended-particle size distributions, and CDOM sources and processing. To equip $g(\cdot)$ with the ability to adapt to different SIOP scenarios, each task $\tau_i$ is sampled in a way that its synthetic samples from $\mathcal{D}$ share similar SIOPs. \hl{The four SIOP spectra ($a_d^*$, $a_y^*$, $a_{ph}^*$, and $b_{bp}^*$) were firstly concatenated, and each wavelength-specific feature was normalised using its mean and sample standard deviation across the dataset. Then,} for each task, we pick a random sample from $\mathcal{D}$, compute its Euclidean distances to all other samples in $\mathcal{D}$ in terms of SIOPs \hl{in the normalised space}, and select the closest $2K_i$ ones. In this way, different tasks correspond to distinct SIOP scenarios.
The adaptation ability of the base model, $g_{\boldsymbol{\theta}}(\cdot)$, is learned by adapting its parameters to each of the sampled tasks. Here, we distinguish the parameters of the base model, $\boldsymbol{\theta}$, which serve as meta-parameters shared across all tasks, from the task-specific parameters, $\boldsymbol{\phi}_i$, which are obtained by adapting $\boldsymbol{\theta}$ to the $i$-th task $\tau_i$. In other words, \(\boldsymbol{\theta}\) represents the general knowledge shared across tasks, 
while \(\boldsymbol{\phi}_i\) captures the additional task-specific knowledge learned from the particular task \(\tau_i\). For the $i$-th task $\tau_i$, \(\boldsymbol{\phi}_i\) are obtained by adapting $\boldsymbol{\theta}$ to the support set $\mathcal{S}_i$, via minimising the following loss function $\mathcal{L}_{\mathcal{S}_i}$ that measures the discrepancy between the model-predicted and the true values of TSS, DOC, and TChl-a:
\begin{equation}
\label{eq:support_loss}
\begin{split}
&\mathcal{L}_{\mathcal{S}_i}(\boldsymbol{\phi}_i)
= \\
&\frac{1}{K_i}
\sum_{\substack{
(R_{rs},\, [\mathrm{TSS},\, \mathrm{DOC},\\
\mathrm{TChl\text{-}a}]) \in \mathcal{S}_i
}}
\big\|
g_{\boldsymbol{\phi}_i}(R_{rs})
-
[\mathrm{TSS},\, \mathrm{DOC},\, \mathrm{TChl\text{-}a}]
\big\|_2^2.
\end{split}
\end{equation}

\hl{For each task $\tau_i$, the inner-loop optimisation was initialised using the shared meta-parameters, $\boldsymbol{\phi}_i^{(0)}=\boldsymbol{\theta}$. The task-specific parameters were subsequently updated on the support set according to:}
\begin{equation}
\colorbox{white}{$\displaystyle
\boldsymbol{\phi}_i^{(u+1)}
=
\boldsymbol{\phi}_i^{(u)}
-
\alpha
\nabla_{\boldsymbol{\phi}_i^{(u)}}
\mathcal{L}_{S_i}\!\left(\boldsymbol{\phi}_i^{(u)}\right),
\quad u=0,\ldots,U-1,
$}
\end{equation}
\hl{where $\alpha=10^{-3}$ is the inner-loop learning rate and $U=200$ is the number of inner-loop gradient updates.}

The accuracy of this adaptation is then measured on the query set with the same type of loss function $\mathcal{L}_{\mathcal{Q}_i}$. Then, the parameters of the base model, $\boldsymbol{\theta}$, are optimised by minimising the meta loss, $\mathcal{J}(\boldsymbol{\theta})$, which is expressed as the average query loss across all tasks:
\begin{equation}
\label{eq:meta_loss}
   \mathcal{J}(\boldsymbol{\theta})=\frac{1}{B}\sum_{i=1}^{B}
\mathcal{L}_{\mathcal{Q}_i}\!\big(\boldsymbol{\phi}_i(\boldsymbol{\theta})\big), 
\end{equation}
where $B$ is \hl{the number of tasks in each meta-batch}, and $\boldsymbol{\phi}_i(\boldsymbol{\theta})$ stands for the adapted parameters $\boldsymbol{\phi}_i$ to the $i$-th task $\tau_i$ under the meta-parameters $\boldsymbol{\theta}$. \hl{Each outer-loop update used $B=64$ tasks, each with $K_i$ support samples and $K_i$ query samples, where $K_i$ was sampled uniformly from the integers between 100 and 1000. The meta-parameters were optimised using Adam with a fixed learning rate of $10^{-4}$ for 101 outer-loop updates, retaining second-order derivatives through the inner-loop updates.}

\subsection{Region-specific model adaptation}
\label{ssec:adaptation}

We adapt the pretrained base model, $g_{\boldsymbol{\theta}}(\cdot)$, to each experimental site using local in situ samples. Let $D_r = \{(R^{(i)}_{rs}, \boldsymbol{y}^{(i)})\}_{i=1}^{N_r}$ denote the region-specific dataset collected from the experimental site $r$, where $R^{(i)}_{rs} \in \mathbb{R}^{M}$ represents the hyperspectral $R_{rs}$ spectrum and $\boldsymbol{y}^{(i)} = [\mathrm{TSS}, \mathrm{DOC}, \mathrm{TChl\text{-}a}] \in \mathbb{R}^{3}$ represents the corresponding \hl{WQ} parameters measured in situ, with $N_r$ denoting the number of samples available for that site. Following the meta-learning paradigm introduced in Subsection \ref{ssec:pretraining}, we treat each site as a new task and initialise the model parameters using the pretrained meta-parameters $\boldsymbol{\theta}$. The region-specific parameters $\boldsymbol{\phi}_r$ are then obtained through iterative gradient updates using the regional dataset. Specifically, for a given set of parameters $\boldsymbol{\phi}_r$, the adaptation loss is defined as:
\begin{equation}
\label{eq:adaptation}
\mathcal{L}_r(\boldsymbol{\phi}_r) =
\frac{1}{N_r}
\sum_{i=1}^{N_r}
\left\|
g_{\boldsymbol{\phi}_r}(R^{(i)}_{rs}) - \boldsymbol{y}^{(i)}
\right\|_2^2 .
\end{equation}
This procedure iteratively adjusts the model parameters so that the predictions align with the local \hl{WQ}--$R_{rs}$ relationships observed in the target experimental site. 

\subsection{Accuracy assessment}

In this study, a 10-fold cross-validation scheme was adopted for accuracy assessment. To mitigate temporal autocorrelation, a block-based training/test strategy was employed in which the dataset from each experimental site was partitioned into 10 temporally separate blocks.

The proposed approach was benchmarked against seven existing models, including three  empirical models, \hl{one semi-analytical model}, and \hl{three} deep-learning-based models. The empirical models are the TSS model by \hl{Choo et al.}~\cite{choo2022spatial}, the DOC model by \hl{Cherukuru et al.}~\cite{cherukuru2016estimating}, and the TChl-a model by \hl{Cherukuru et al.}~\cite{cherukuru2019bio}. \hl{The semi-analytical model is the quasi-analytical algorithm (QAA) (Version 6)~\cite{Lee2002,IOCCG2015QAAv6}. As QAA retrieves optical properties rather than mass concentrations, TSS was estimated from the retrieved $b_{bp}(\lambda_{555})$ using an empirically calibrated power-law relationship, $\mathrm{TSS}=\alpha[b_{bp}(\lambda_{555})]^\beta$.} The deep-learning-based models include \hl{a simple multi-layer perceptron (MLP) network}, the DL-RS model~\cite{unnithan2025}, which was used for benchmarking TSS and DOC, and the HyperEST model~\cite{LUO2025104761}, which was used for benchmarking TSS and TChl-a. These benchmark models were recalibrated/fine-tuned for each experimental site using our dataset. Consistent with the proposed method, a temporal block-based 10-fold cross-validation scheme was employed to evaluate model performance.

Log-scaled metrics have been recommended for accuracy assessment in literature mainly because \hl{WQ} measurements are often skewed towards small values. Following the practice by \hl{Unnithan et al.}~\cite{unnithan2025}, \hl{Luo et al.}~\cite{LUO2025104761}, and \hl{Seegers et al.}~\cite{seegers2018performance}, we adopted logarithmic coefficient of determination (Log-$R^2$) , logarithmic mean absolute error (Log-MAE), \hl{logarithmic bias (Log-Bias)} as assessment metrics:
\begin{equation}
\label{eq:log_r2}
\mathrm{Log}\text{-}R^{2}
=
1-
\frac{
\sum_{i=1}^{n}
\left[
\log_{10}(y_{\mathrm{true},i})
-
\log_{10}(y_{\mathrm{pred},i})
\right]^{2}
}{
\sum_{i=1}^{n}
\left[
\log_{10}(y_{\mathrm{true},i})
-
\overline{\log_{10}(y_{\mathrm{true}})}
\right]^{2}
},
\end{equation}
\begin{equation}
\mathrm{Log}\text{-}\mathrm{MAE}
=
10^{
\frac{1}{n}
\sum_{i=1}^{n}
\left|
\log_{10}(y_{\mathrm{pred},i})
-
\log_{10}(y_{\mathrm{true},i})
\right|
}
-1,
\end{equation}
\begin{equation}
\mbox{\hl{$\displaystyle
\mathrm{Log\text{-}Bias}
=
10^{\frac{1}{n}\sum_{i=1}^{n}
\log_{10}\left(\frac{y_{\mathrm{pred},i}}{y_{\mathrm{true},i}}\right)}
-1.
$}}
\end{equation}
\hl{To distinguish within-site retrieval performance from between-site concentration differences, Log-$R^2$ was calculated for each experimental site using Eq.~(\ref{eq:log_r2}), and the site-mean Log-$R^2$ was then reported as the mean of the five site-specific Log-$R^2$ values.}
\section{Results and Discussion}
\label{sec:results_and_discussions}

\subsection{Overview}


The following Subsections are organised as follows. Subsections \ref{ssec:analysis_of_the_spectral_library} to \ref{ssec:analysis_of_the_synthetic_dataset} show the analysis results of the spectral library, forward model, and generated dataset. Subsection \ref{ssec:analysis_of_regional_distinctions} analyses the regional differences in \hl{WQ} parameters and hyperspectral $R_{rs}$ among the experimental sites. Subsection \ref{ssec:retrieval_accuracies} shows the \hl{WQ} retrieval accuracies obtained with the proposed approach, in comparison with the benchmark models. Subsection \ref{ssec:timeseries_analysis} analyses the time-series variations in $R_{rs}$ and \hl{WQ} parameters and compares the predicted values with in situ measurements. \hl{Subsection \ref{ssec:noise-sensitivity} assesses retrieval sensitivity to additional $R_{rs}$ noise. Subsection \ref{ssec:pace-application} presents an application to PACE OCI observations. Subsection \ref{ssec:comparison_with_existing_methods} discusses the comparison of the proposed approach with existing methods. Finally, Subsection \ref{ssec: limitations_and_future_work} discusses the limitations and future work.}


\subsection{Analysis of the bio-optical spectral library}
\label{ssec:analysis_of_the_spectral_library}

The statistics of the spectral library samples are summarised in Table~\ref{tab:sl}. As shown in the table, the spectral library covers a broad dynamic range for each parameter. In particular, the samples extend from near-freshwater conditions (0.08) to fully marine salinities (39.4), and from relatively cool (12.1~\textdegree{}C) to warm waters (31.4~\textdegree{}C). This indicates that the spectral library encompasses a wide range of environmental conditions representative of both river-affected estuarine and open-ocean waters. The broad representativeness of the spectral library samples provides a basis to guide the pretraining of the base model, allowing it to learn generalised \hl{WQ}--$R_{rs}$ relationships across diverse water types and environmental conditions.

\begin{table}[!t]
\centering
\caption{Statistics of the \hl{WQ} parameters, specific inherent optical properties, and ancillary physical variables in the bio-optical spectral library.}
\label{tab:sl}

\begin{tabularx}{\columnwidth}{
    >{\raggedright\arraybackslash}p{2.8cm}
    *{3}{>{\centering\arraybackslash}X}
    >{\centering\arraybackslash}p{2.2cm}
}
\toprule
Parameter & Min & Med & Max & Mean $\pm$ Std Dev \\
\midrule

$T$ (\textdegree{}C)
& 12.128 & 25.190 & 31.404 & $24.301 \pm 5.137$ \\

$S$ (--)
& 0.080 & 33.420 & 39.400 & $28.388 \pm 9.851$ \\

TSS (mg/L)
& 0.133 & 2.600 & 69.830 & $6.892 \pm 10.637$ \\

DOC (mg/L)
& 0.240 & 1.300 & 14.250 & $2.177 \pm 2.200$ \\

TChl-a ($\mu$g/L)
& 0.059 & 0.963 & 22.037 & $2.204 \pm 3.330$ \\

$a^*_{y}(\lambda_{440})$ (m$^2$ g$^{-1}$)
& 0.004 & 0.230 & 2.552 & $0.307 \pm 0.288$ \\

$b^*_{bp}(\lambda_{550})$ (m$^2$ g$^{-1}$)
& 0.001 & 0.008 & 0.042 & $0.008 \pm 0.006$ \\

$a^*_{d}(\lambda_{440})$ (m$^2$ g$^{-1}$)
& 0.002 & 0.031 & 0.257 & $0.037 \pm 0.031$ \\

$a^*_{ph}(\lambda_{440})$ (m$^2$ mg$^{-1}$)
& 0.010 & 0.059 & 0.400 & $0.076 \pm 0.060$ \\

\bottomrule
\end{tabularx}
\end{table}

\begin{figure}[!t]
\centering
\includegraphics[width=\columnwidth]{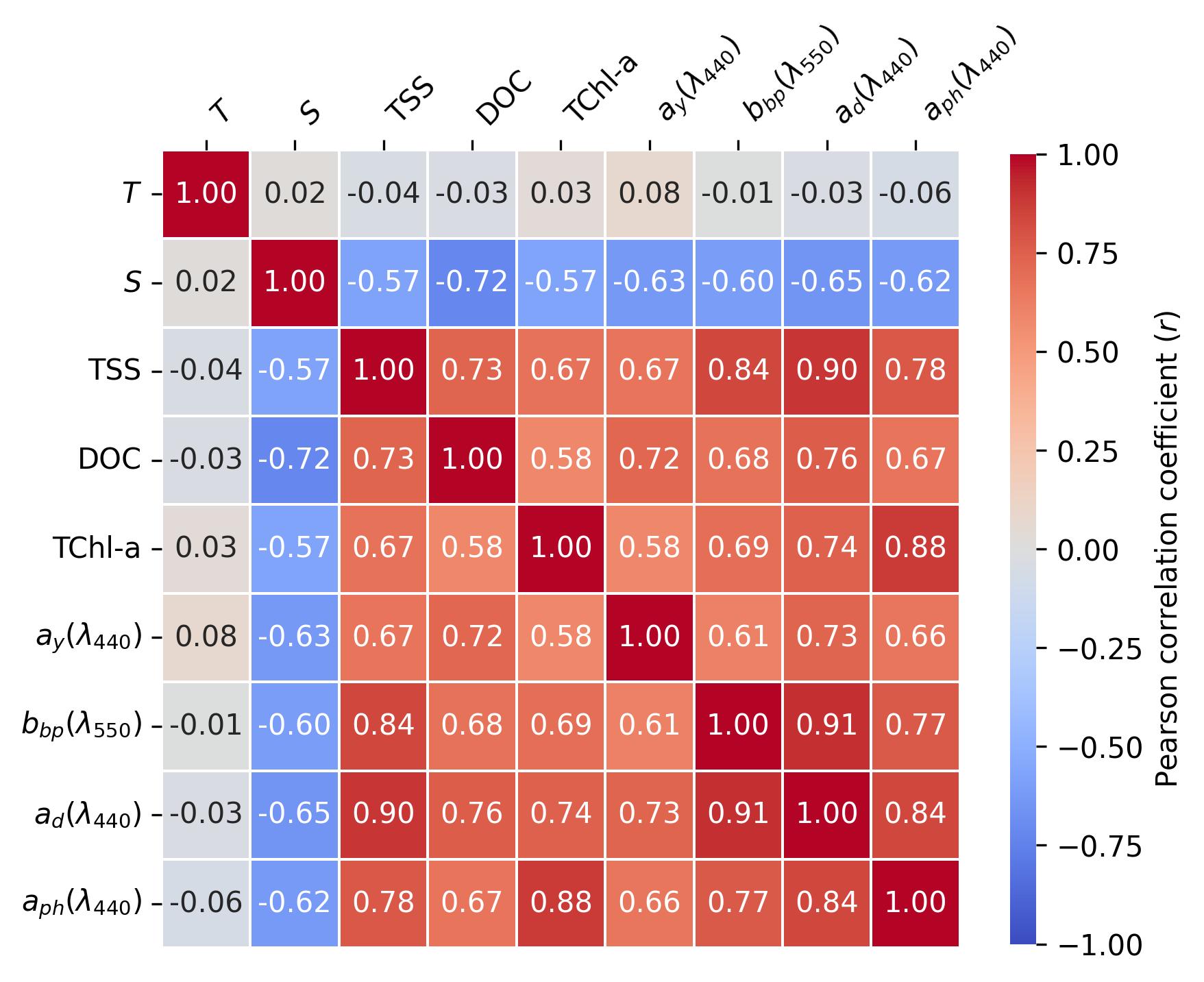}
\caption{Pearson correlation matrix among \hl{WQ} parameters (TSS, DOC, and TChl-a), IOPs ($a_{y}(\lambda_{440})$, $b_{bp}(\lambda_{550})$, $a_{d}(\lambda_{440})$, and $a_{ph}(\lambda_{440})$) and ancillary physical variables ($T$ and $S$) for samples in the bio-optical spectral library.
\label{fig:matrix}}
\end{figure}

Fig. \ref{fig:matrix} presents the Pearson correlation matrix for the spectral library measurements, where the coefficients of correlation ($r$) were computed for pairwise combinations of $T$, $S$, TSS, DOC, TChl-a, $a_{y}(\lambda_{440})$, $b_{bp}(\lambda_{550})$, $a_{d}(\lambda_{440})$, and $a_{ph}(\lambda_{440})$. It was found that water temperature $T$ exhibits weak correlations with \hl{WQ} parameters and IOPs with $|r|\leq 0.08$, which is expected since water constituents are primarily governed by hydrological inputs and transport processes rather than by temperature itself. In contrast, water salinity $\mathit{S}$ is generally negatively correlated with \hl{WQ} parameters and IOPs with $r$ ranging from \textminus 0.72 to \textminus 0.57, as freshwater inflows reduce seawater salinity and concurrently introduce elevated loads of particulates and dissolved matter from terrestrial sources. TSS exhibits strong correlations with particulate IOPs: it is highly correlated with $b_{bp}(\lambda_{550})$ ($r=0.84$), $a_{d}(\lambda_{440})$ ($r=0.90$), and $a_{ph}(\lambda_{440})$ ($r=0.78$). This is physically consistent, as TSS represents the concentration of suspended particulate matter, encompassing both inorganic NAP and organic phytoplankton components, and therefore influences both light backscattering and absorption. DOC shows a strong positive correlation with $a_{y}(\lambda_{440})$ ($r=0.72$). This relationship is expected because DOC is closely associated with CDOM concentration, which highly affects light absorption in the blue spectral region. TChl-a shows reasonably strong correlation with $a_{ph}(440)$ ($r=0.88$), reflecting the dependence of phytoplankton absorption on chlorophyll concentration. The results of this correlation analysis indicate that the underlying physical correlations among variables are well preserved in the spectral library measurements.

\begin{figure}[!t]
\centering
\includegraphics[width=\columnwidth]{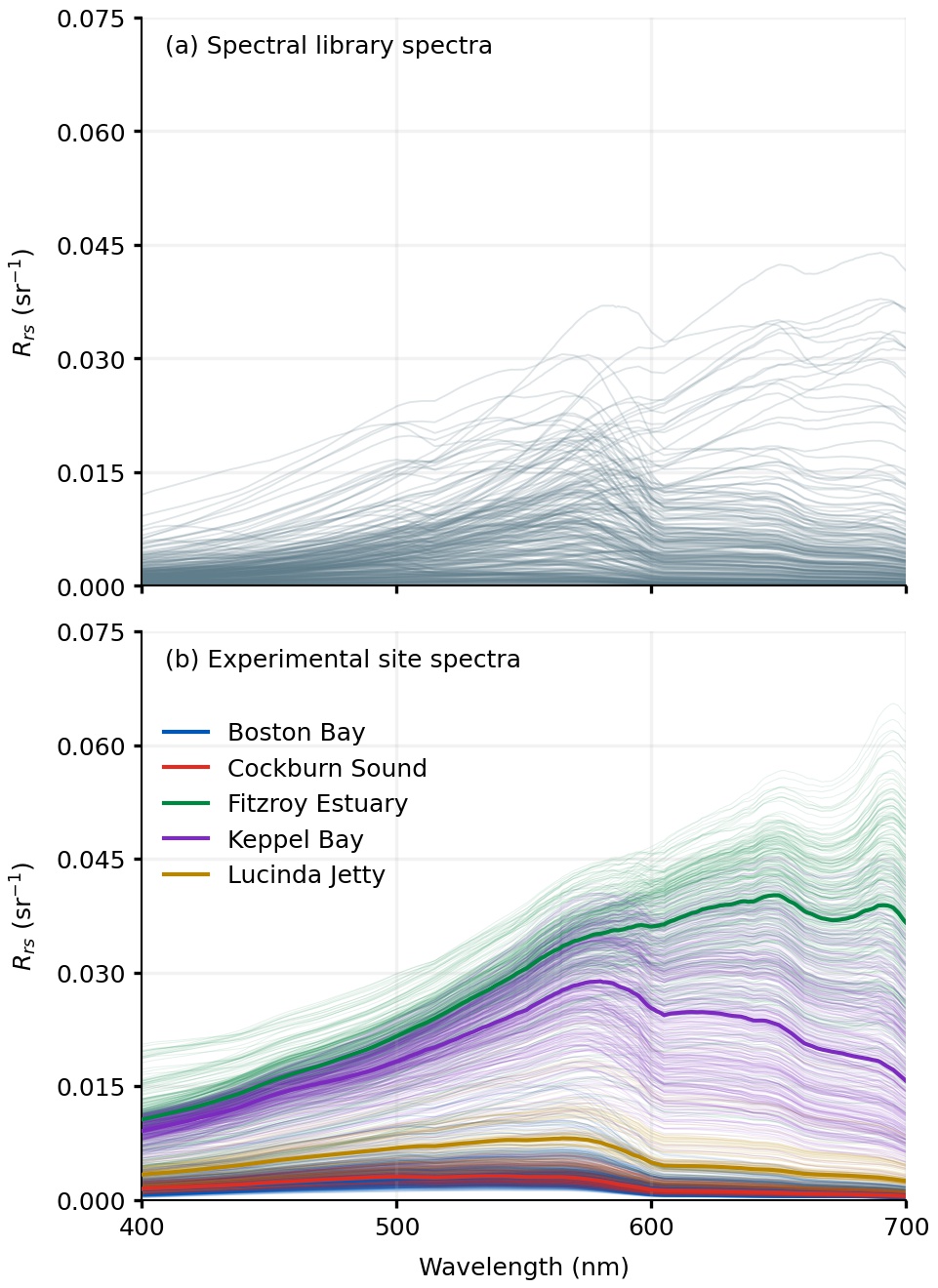}
\caption{\hl{(a) Forward-modelled $R_{rs}$ spectra from the bio-optical spectral library. (b) Example $R_{rs}$ spectra from the experimental sites (thick lines indicate the median $R_{rs}$ spectrum at each site; thin lines show individual spectra).}
\label{fig:spectra}}
\end{figure}

\hl{Fig.~\ref{fig:spectra} compares forward-modelled $R_{rs}$ spectra from the spectral library with in situ measured spectra from the five experimental sites. This comparison provides qualitative context for the spectral shapes and magnitudes represented by the two datasets. It is worth noting that Fig.~\ref{fig:spectra} does not independently validate the forward model, as the two datasets do not contain matched water samples.}

\begin{figure}[!t]
\centering
\includegraphics[width=0.85\columnwidth]{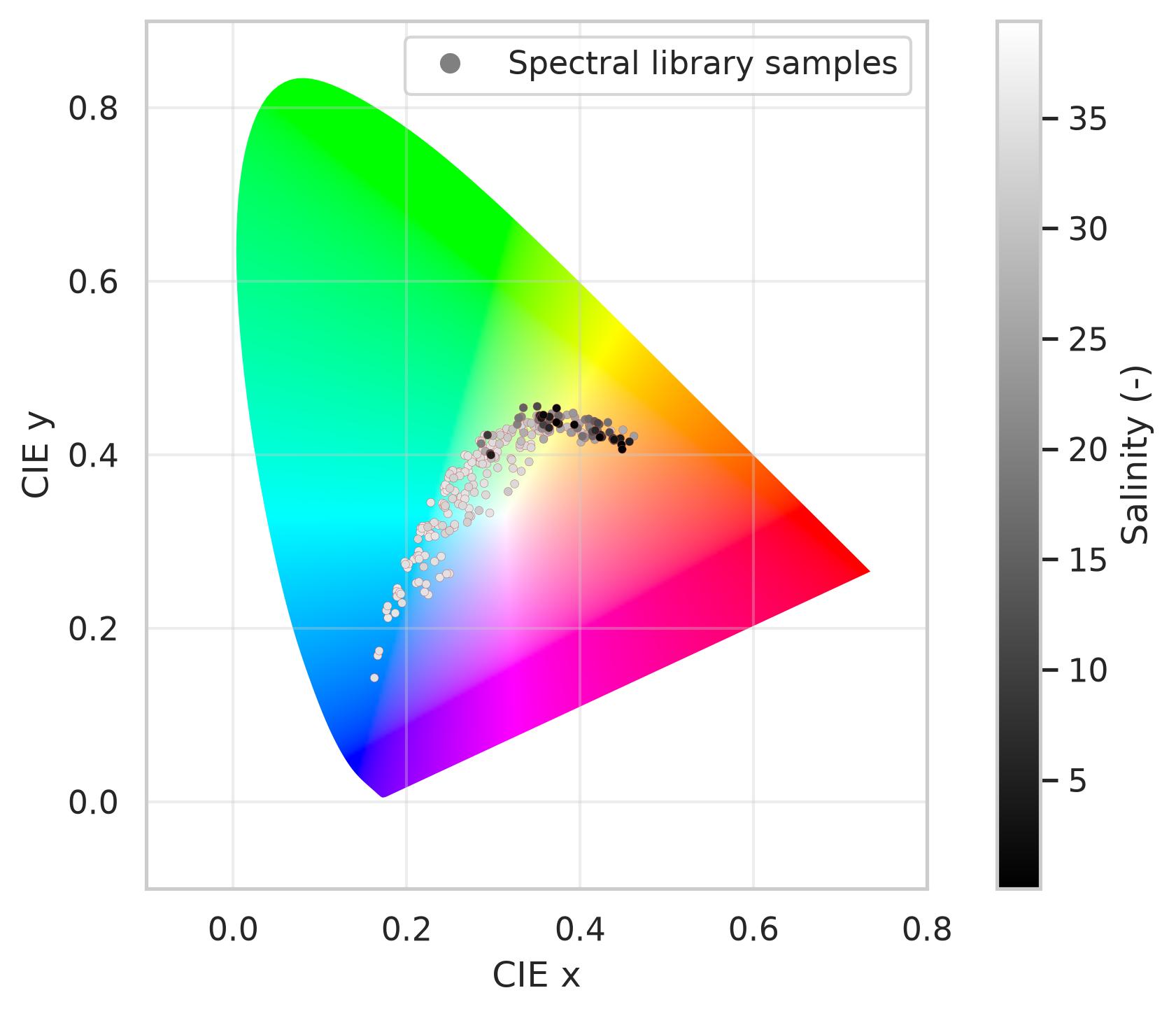}
\caption{The CIE 1931 chromaticity diagram showing the colour distribution of spectral library samples. Sample chromaticity coordinates were computed from $R_{rs}$ spectra using the CIE 1931 2\textdegree{} Standard Observer under CIE Standard Daylight D65 illumination. Sample points are shaded in grayscale according to their salinity. \label{fig:owt}}
\end{figure}

The CIE 1931 chromaticity diagram in Fig.~\ref{fig:owt} shows the colour distribution of in situ samples in the bio-optical spectral library. The distribution of these samples follows a trajectory that closely resembles the chromaticity progression of the Forel–Ule (FU) colour scale (\emph{e.g.}, \cite{wernand2010spectral,novoa2013forel}), and spans the FU range from blue and green to yellow and brown waters. It was also observed from that figure that salinity exhibits a systematic gradient along the chromaticity trajectory, with higher-salinity samples generally associated with greener to bluer colours, and lower-salinity samples tending towards yellower to browner colours. These findings indicate that the bio-optical spectral library employed in this study provides a representative coverage of diverse water colours.

\subsection{Analysis of the bio-optical forward model}
\label{ssec:analysis_of_the_forward_model}

\begin{figure}[!t]
\centering
\includegraphics[width=0.95\columnwidth]{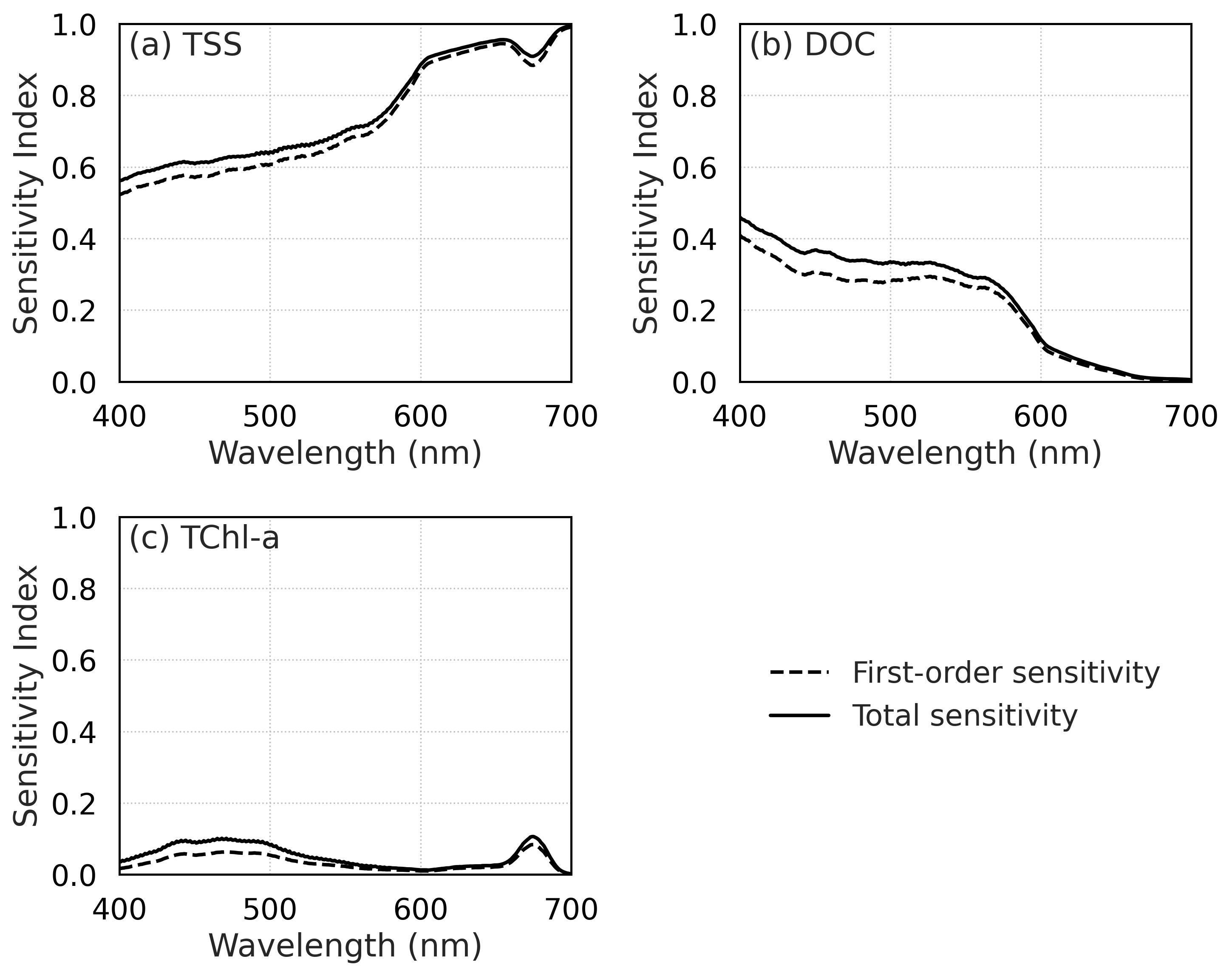}
\caption{Sensitivity indices of $R_{rs}$ to variations in (a) TSS, (b) DOC, and (c) TChl-a across the 400--700~nm spectral range.
}
\label{fig:sensitivity}
\end{figure}

Fig.~\ref{fig:sensitivity} shows the sensitivity indices of the bio-optical forward model as computed with the Extended Fourier Amplitude Sensitivity Test (EFAST)~\cite{saltelli1999quantitative}. It was found that the sensitivity of $R_{rs}$ to TSS increases toward longer wavelengths, reflecting the higher influence of particle backscattering in the red region. In contrast, the sensitivity of $R_{rs}$ to DOC is relatively higher in the blue region than in the red region, which is consistent with the relatively higher absorption of CDOM at shorter wavelengths. Compared with TSS and DOC, the sensitivity of $R_{rs}$ to TChl-a displays overall lower sensitivity, with relatively higher sensitivity in the blue and near 680 nm regions corresponding to known chlorophyll absorption features in these regions. Previous studies have shown that absorption is generally more pronounced in the blue region than in the red due to CDOM and phytoplankton pigments, whereas scattering effects associated with suspended particles tend to become relatively more important at longer wavelengths (\emph{e.g.},~\cite{manzo2015sensitivity}). Compared with these well-established spectral characteristics of \hl{WQ} parameters, the sensitivity analysis results in Fig.~\ref{fig:sensitivity} suggest that the forward model is built upon assumptions that align well with the bio-optical properties of water constituents.

\subsection{Analysis of the synthetic dataset}
\label{ssec:analysis_of_the_synthetic_dataset}

\begin{figure*}[!t]
\centering
\includegraphics[width=\textwidth]{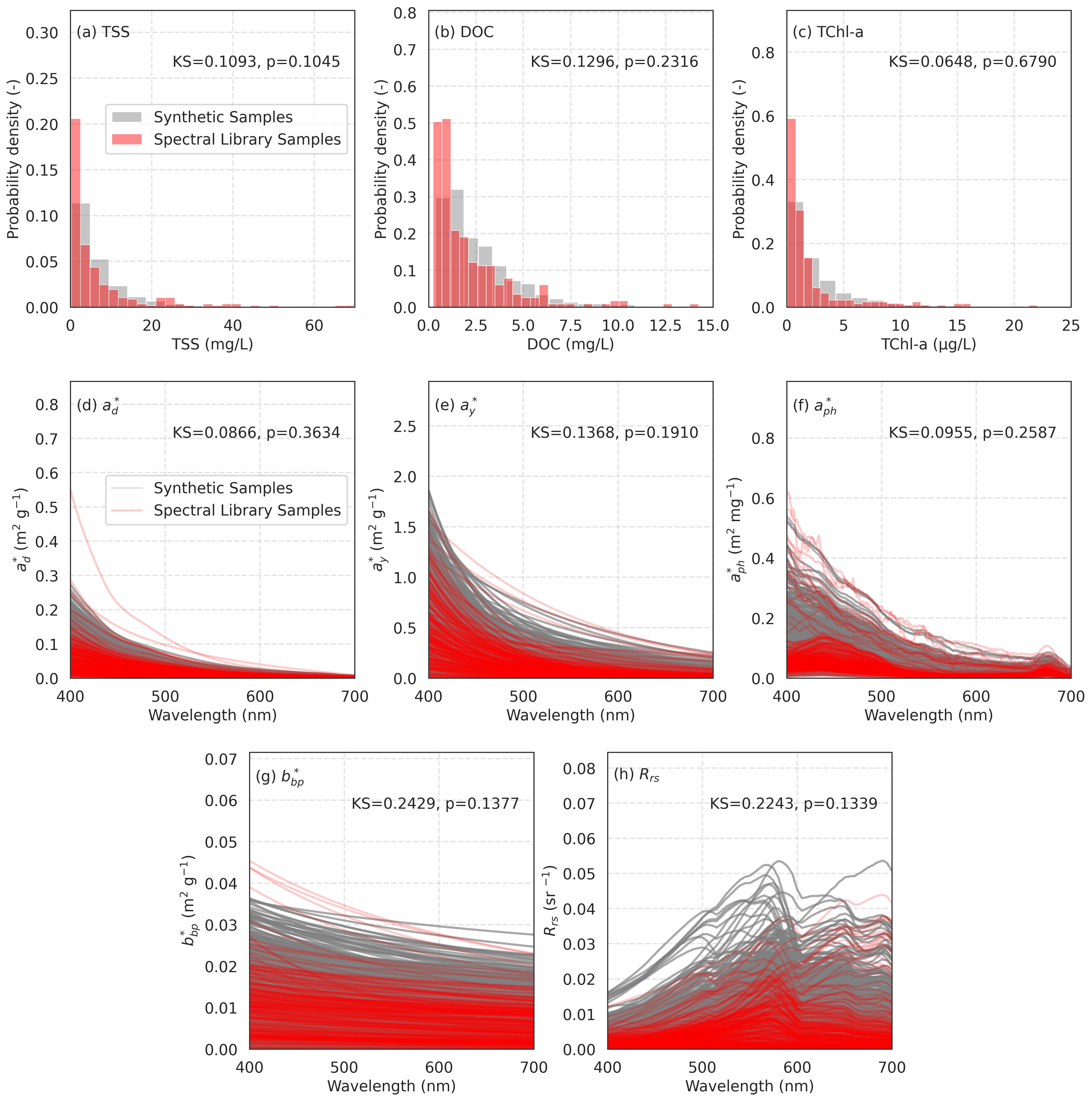}
\caption{Comparisons of samples from the synthetic dataset $\mathcal{D}$ with those from the in situ spectral library for (a--c) \hl{WQ} parameters (TSS, DOC, and TChl-a), (d--g) SIOPs ($a^*_d$, $a^*_y$, $a^*_{ph}$, and $b^*_{bp}$), and (h) the corresponding $R_{rs}$. The Wasserstein distance is calculated to quantify the distributional difference. \label{fig:distribution}}
\end{figure*}

Fig.~\ref{fig:distribution} shows the distribution of 10,000 samples from the synthetic dataset $\mathcal{D}$ in comparison with samples from the in situ spectral library. The histogram comparisons in Figs.~\ref{fig:distribution}a--c show that the \hl{WQ} parameters (TSS, DOC, and TChl-a) in the synthetic dataset not only span a similar range of values as the spectral library measurements, but also generally follow their distributional patterns. The Wasserstein distance between the synthetic and spectral library distributions is 0.53 mg/L for TSS, 0.12 mg/L for DOC, and 0.35 $\mu$g/L for TChl-a. The spectral comparisons in Figs.~\ref{fig:distribution}d–g indicate that the synthetic SIOP spectra ($a^*_d$, $a^*_y$, $a^*_{ph}$, and $b^*_{bp}$) largely preserve the spectral correlations across wavelengths (\emph{e.g.},~spectral smoothness and absorption features) observed in the spectral library measurements, and their shapes and magnitudes also generally resemble those of the measured spectra. The spectrally averaged Wasserstein distance is 0.0059 m\textsuperscript{2} g\textsuperscript{\textminus 1} for $a^*_d$, 0.0505 m\textsuperscript{2} g\textsuperscript{\textminus 1} for $a^*_y$, 0.0051 m\textsuperscript{2} mg\textsuperscript{\textminus 1} for $a^*_{ph}$, and 0.0016 m\textsuperscript{2} g\textsuperscript{\textminus 1} for $b^*_{bp}$. In particular, the chlorophyll-a absorption peaks in $a^*_{ph}$ at 430–440 nm and 665–675 nm, evident in the spectral library spectra, are well preserved in the synthetic spectra, as shown from Fig.~\ref{fig:distribution}f. The simulated $R_{rs}$ spectra also closely resemble those from the spectral library, in terms of both spectral shape and magnitude, as shown in Fig.~\ref{fig:distribution}h, with a spectrally averaged Wasserstein distance of 0.0009 sr\textsuperscript{\textminus 1}. The close agreement between the distributions of the synthetic samples and the spectral library samples, as shown in Fig.~\ref{fig:distribution}, demonstrates that the synthetic dataset \hl{preserves the distributional characteristics of the spectral library samples}.

\begin{figure*}[!t]
\centering
\includegraphics[width=\textwidth]{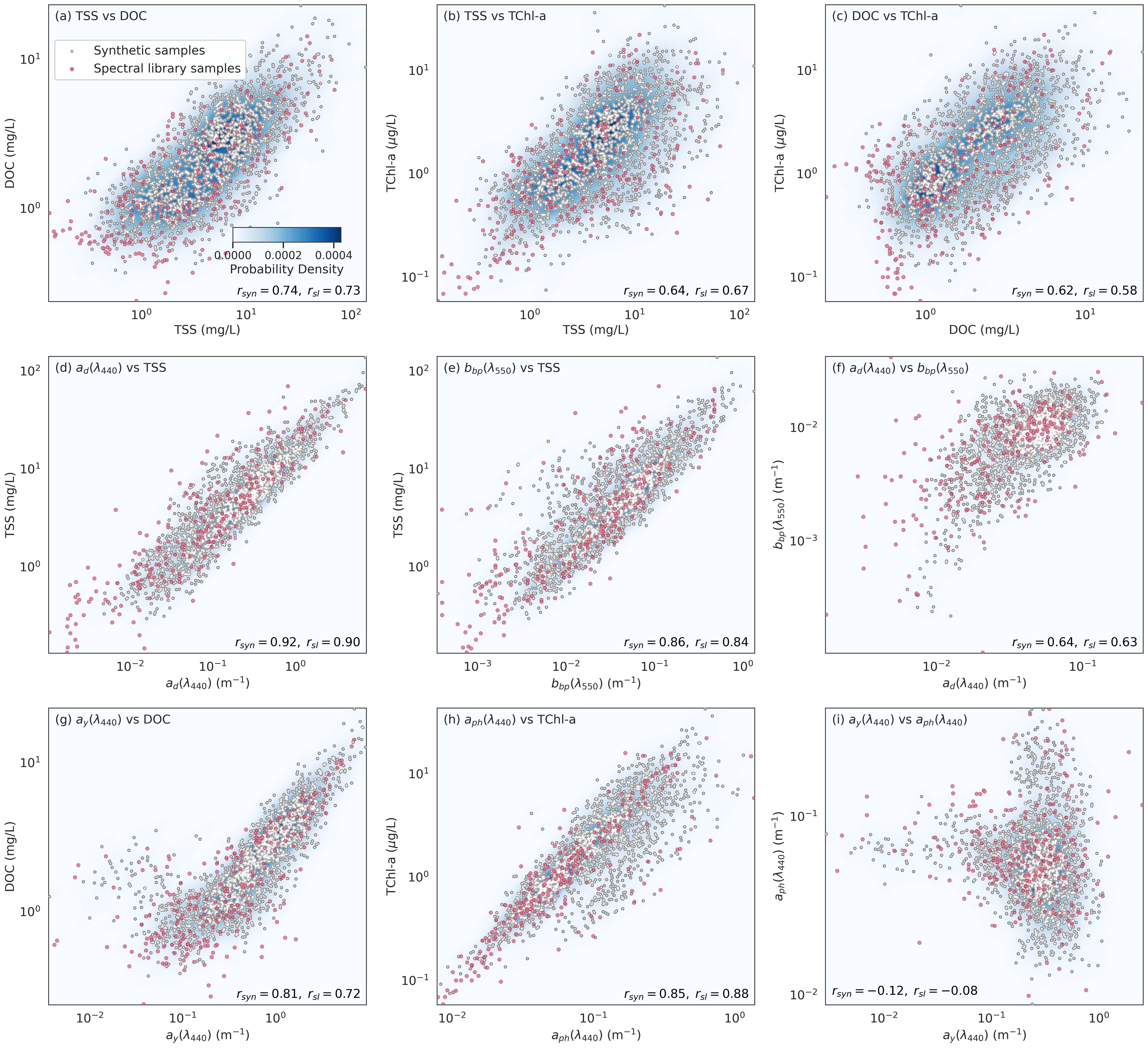}
\caption{Comparison of parameter correlations between synthetic samples and spectral library samples for the joint distributions of (a) TSS and DOC, (b) TSS and TChl-a, (c) DOC and TChl-a, (d) \hl{$a_d(\lambda_{440})$ and TSS}, (e) \hl{$b_{bp}(\lambda_{550})$ and TSS}, (f) $a_d(\lambda_{440})$ and $b_{bp}(\lambda_{550})$, (g) \hl{$a_y(\lambda_{440})$ and DOC}, (h) \hl{$a_{ph}(\lambda_{440})$ and TChl-a}, and (i) $a_y(\lambda_{440})$ and $a_{ph}(\lambda_{440})$. Background colour indicates the probability density of the joint distributions. $r_{syn}$ and $r_{sl}$ stand for the Pearson correlation coefficients between synthetic samples and spectral library samples, respectively. \label{fig:correlation}}
\end{figure*}

While Fig.~\ref{fig:distribution} presents distributional comparisons between the synthetic and spectral library samples for individual parameters, we further assessed whether the inter-parameter correlations observed in the spectral library are preserved in the synthetic dataset. Fig.~\ref{fig:correlation} presents the joint distributions of selected parameter pairs for both synthetic and spectral library samples. The figure shows that the correlation patterns between parameters are similar for the synthetic and spectral library samples. A comparison of Pearson correlation coefficients ($r$) indicates that, for each parameter pair, the $r$ values between the spectral library and synthetic data are closely aligned, with differences ranging from 0.01 (Figs.~\ref{fig:correlation}a and f) to 0.09 (Fig.~\ref{fig:correlation}g). The results shown in Fig.~\ref{fig:correlation} indicate that the inter-parameter correlations observed in the spectral library have been largely preserved in the synthetic dataset.

\subsection{Analysis of regional distinctions}
\label{ssec:analysis_of_regional_distinctions}

\begin{figure}[!t]
\centering
\includegraphics[width=0.9\columnwidth]{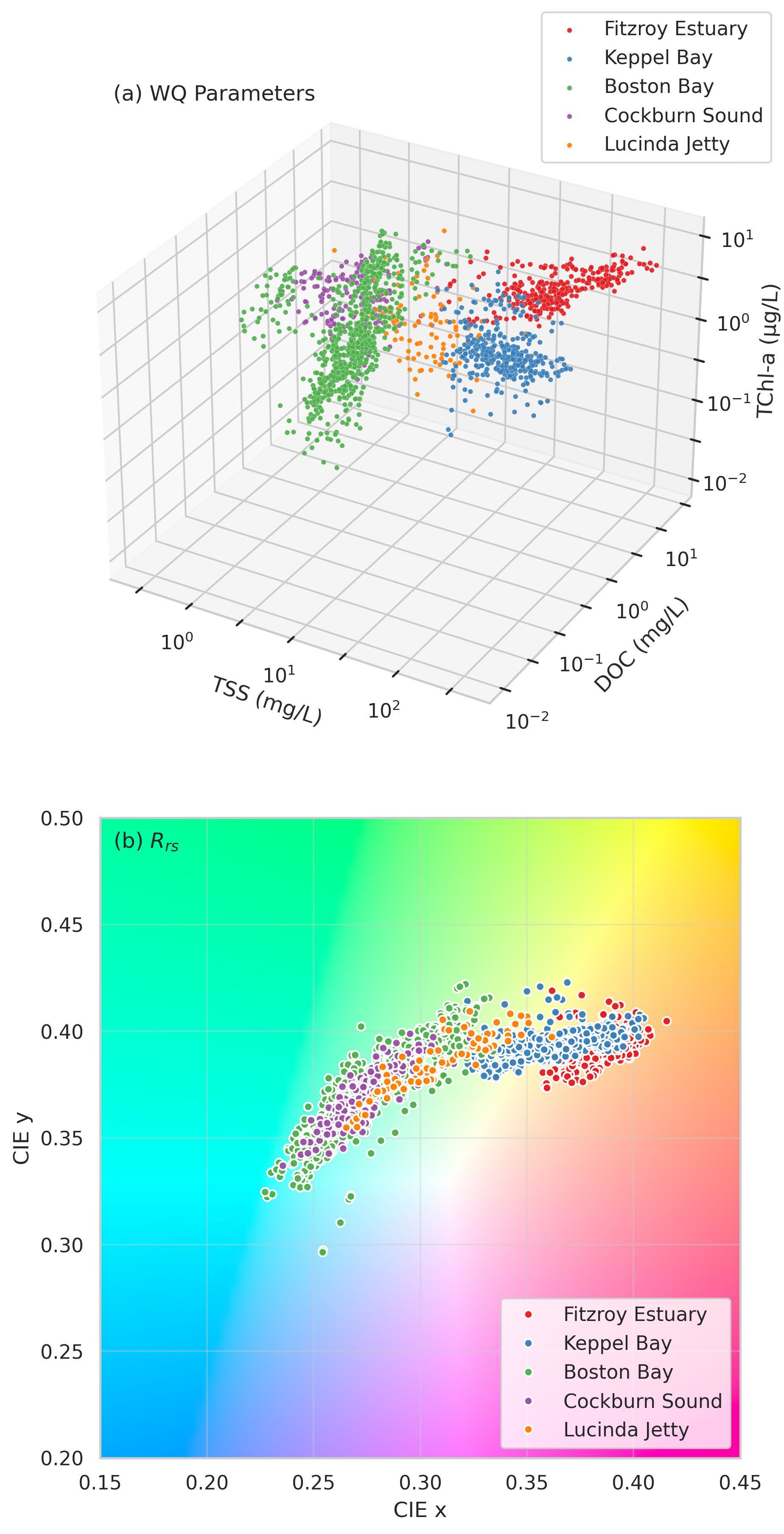}
\caption{(a) Distribution of \hl{WQ} parameters, including TSS, DOC, and TChl-a concentrations, at the experimental sites. (b) The CIE 1931 chromaticity diagram showing the distribution of water-leaving $R_{rs}$ measured at the experimental sites.}
\label{fig:cluster}
\end{figure}

Fig. \ref{fig:cluster} shows the regional distinctions among the experimental sites in their \hl{WQ} and $R_{rs}$ distributions. Fig. \ref{fig:cluster}a shows the distribution of TSS, DOC, and TChl-a concentrations in the \hl{WQ} feature space across the five experimental sites. Fitzroy Estuary exhibits the highest TSS and DOC concentrations overall, with moderate to elevated TChl-a levels. Keppel Bay is characterised by moderate TSS and DOC concentrations but comparatively lower TChl-a values. Boston Bay and Cockburn Sound cluster towards lower TSS and DOC concentrations, with Boston Bay showing a broader spread in TChl-a than Cockburn Sound. Lucinda Jetty shows moderate TSS, DOC, and TChl-a values as compared with other sites.

In the CIE 1931 chromaticity diagram shown in Fig. \ref{fig:cluster}b, the $R_{rs}$ measurements display distinct colour groupings that reflect differences in apparent water colour among the regions. Waters from Fitzroy Estuary display brown colours, characteristic of turbid estuarine conditions. Keppel Bay exhibits less brown colours due to moderate river influence. Boston Bay and Cockburn Sound are associated with clearer green to blue–green waters, typical of less turbid coastal environments with lower suspended sediment loads. Lucinda Jetty is more green-coloured than the clearer coastal sites but less brown than the estuarine waters.

In summary, the regional distinctions observed in Fig.~\ref{fig:cluster} are clearly reflected in both the \hl{WQ} parameters and their corresponding optical responses across the experimental sites. This variability suggests that constituent dynamics and $R_{rs}$ characteristics are region-dependent. These findings underpin the need to develop region-adaptable approaches for \hl{WQ} parameter retrieval to ensure robust, accurate, and transferable performance across diverse coastal and estuarine water environments.

\subsection{Accuracies of \hl{WQ} parameter retrieval}
\label{ssec:retrieval_accuracies}

\begin{figure*}[!t]
\centering
\includegraphics[width=0.75\textwidth]{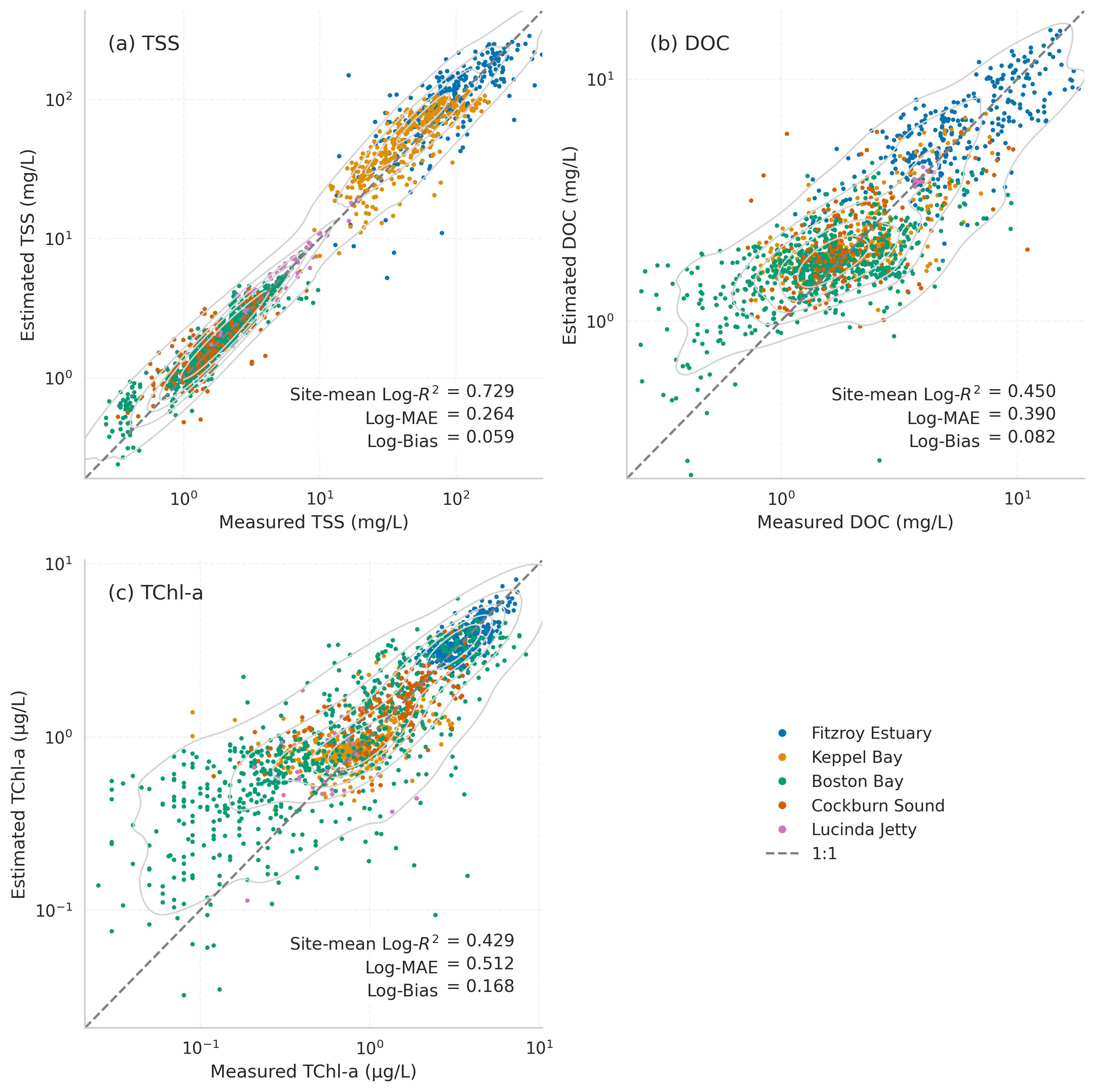}
\caption{Comparison between predicted and measured concentrations of (a) TSS, (b) DOC, and (c) TChl-a across the five experimental sites. Axes are shown on $\log_{10}$ scales. Grey lines represent contours of equal probability mass for the purpose of enhancing visualisation of the data distribution. \label{fig:accuracy}}
\end{figure*}

Fig.~\ref{fig:accuracy} shows the relationship between TSS, DOC, and TChl-a measurements and those predicted from $R_{rs}$ using the proposed approach. Across the five experimental sites, TSS showed a \hl{site-mean} Log-$R^2$ of \hl{0.729} and a Log-MAE of 0.264. DOC showed a \hl{site-mean} Log-$R^2$ of \hl{0.450} and a Log-MAE of 0.390. TChl-a showed a \hl{site-mean} Log-$R^2$ of \hl{0.429} and a Log-MAE of 0.512. \hl{Table~\ref{tab:proposed_concentration_stratified} shows the retrieval accuracies for low, medium, and high concentration groups defined by the 1/3 and 2/3 quantiles of measured concentrations for each parameter. It was observed from the table that TChl-a is overestimated at low concentrations and slightly underestimated at high concentrations, with a similar pattern for DOC. Low-concentration TChl-a retrieval may be affected by weak pigment signals and overlapping absorption by other constituents, including DOC-associated CDOM at shorter wavelengths, as shown in the sensitivity patterns in Fig.~\ref{fig:sensitivity}. When spectral information is insufficient to distinguish concentrations, mean-squared-error training may favour predictions closer to typical training values, contributing to overestimation at low concentrations and underestimation at high concentrations.}

\begin{table}[!t]
\centering
\caption{\hl{Concentration-stratified accuracy of the proposed approach.}}
\label{tab:proposed_concentration_stratified}
\small
\setlength{\tabcolsep}{3pt}
\begin{tabular*}{\columnwidth}{@{\extracolsep{\fill}}ll@{\extracolsep{0pt}\hspace{4pt}}r@{\extracolsep{\fill}\hspace{6pt}}r@{}}
\toprule
\hl{Parameter} & \hl{Level} & \hl{Log-MAE} & \hl{Log-Bias} \\
\midrule
\hl{TSS} & \hl{Low (0.3--2.0 mg/L)} & \hl{0.233} & \hl{0.086} \\
 & \hl{Medium (2.0--28.4 mg/L)} & \hl{0.223} & \hl{0.039} \\
 & \hl{High (28.4--428.8 mg/L)} & \hl{0.339} & \hl{0.052} \\
\midrule
\hl{DOC} & \hl{Low (0.2--1.6 mg/L)} & \hl{0.519} & \hl{0.484} \\
 & \hl{Medium (1.6--2.7 mg/L)} & \hl{0.250} & \hl{0.007} \\
 & \hl{High (2.7--25.9 mg/L)} & \hl{0.414} & \hl{-0.152} \\
\midrule
\hl{TChl-a} & \hl{Low (0.1--0.8 $\mu$g/L)} & \hl{0.941} & \hl{0.822} \\
 & \hl{Medium (0.8--1.8 $\mu$g/L)} & \hl{0.359} & \hl{-0.024} \\
 & \hl{High (1.8--8.4 $\mu$g/L)} & \hl{0.313} & \hl{-0.101} \\
\bottomrule
\end{tabular*}
\end{table}

The per-site accuracies are given in Table~\ref{tab:site_accuracy}, with Log-$R^2$ ranging from 0.194 to 0.895 and Log-MAE from 0.014 to 0.792 across sites and parameters. \hl{The differences in retrieval performance among parameters, as seen in Table~\ref{tab:site_accuracy}, may reflect variations in optical sensitivities and bio-optical properties. The stronger TSS retrieval performance is consistent with the greater sensitivity of $R_{rs}$ to TSS than to DOC and TChl-a (Fig.~\ref{fig:sensitivity}), particularly at longer wavelengths where particle backscattering strongly influences reflectance. DOC retrieval is less direct because visible reflectance responds to CDOM absorption rather than the entire DOC pool, and the DOC--CDOM relationship can vary across locations and over time. TChl-a retrieval may be limited by weak pigment absorption signals and their spectral overlap with absorption by CDOM and non-algal particles. Together, these effects contribute to optical non-uniqueness, whereby different constituent combinations produce similar spectra. Site-dependent performance varies by parameter (Table~\ref{tab:site_accuracy}). Weak pigment signals and DOC--CDOM variability may contribute to the low TChl-a Log-$R^2$ at Keppel Bay and DOC Log-$R^2$ at Cockburn Sound, respectively. At Boston Bay, the large TChl-a Log-MAE is consistent with overestimation at low concentrations (Fig.~\ref{fig:accuracy}c). Radiometric noise, surface-reflection errors remaining after correction, and uncertainty in reference measurements may further affect retrieval accuracy. These uncertainties may affect different model classes: purely data-driven models depend on representative training relationships, analytical models on appropriate optical parameterisations, and physics-aware models such as DL-RS~\cite{unnithan2025} and OCRT-Net~\cite{cao2026ocrt} on the adequacy of their physical assumptions and training data. The proposed approach may mitigate some of the region-dependent uncertainties through region-specific adaptation, although it cannot fully resolve all of the aforementioned challenges.}


\begin{table}[!t]
\centering
\footnotesize
\setlength{\tabcolsep}{3pt}
\caption{Per-site accuracies of \hl{water quality} parameters predicted with the proposed approach.}
\label{tab:site_accuracy}

\begin{tabularx}{\columnwidth}{
    >{\raggedright\arraybackslash}p{2cm}
    >{\raggedright\arraybackslash}p{1.2cm}
    >{\centering\arraybackslash}X
    >{\centering\arraybackslash}X
    >{\centering\arraybackslash}X
}
\toprule
Site & Parameter & Log-$R^2$ & Log-MAE & \hl{Log-Bias} \\
\midrule
Fitzroy Estuary & TSS    & 0.539 & 0.390 & \hl{0.078} \\
                & DOC    & 0.515 & 0.375 & \hl{0.087} \\
                & TChl-a & 0.523 & 0.184 & \hl{0.018} \\
Keppel Bay      & TSS    & 0.673 & 0.335 & \hl{0.095} \\
                & DOC    & 0.458 & 0.296 & \hl{0.040} \\
                & TChl-a & 0.194 & 0.372 & \hl{0.063} \\
Boston Bay      & TSS    & 0.869 & 0.186 & \hl{0.034} \\
                & DOC    & 0.349 & 0.462 & \hl{0.104} \\
                & TChl-a & 0.515 & 0.792 & \hl{0.323} \\
Cockburn Sound  & TSS    & 0.668 & 0.296 & \hl{0.059} \\
                & DOC    & 0.272 & 0.386 & \hl{0.082} \\
                & TChl-a & 0.437 & 0.436 & \hl{0.115} \\
Lucinda Jetty   & TSS    & 0.895 & 0.174 & \hl{0.062} \\
                & DOC    & 0.654 & 0.014 & \hl{0.001} \\
                & TChl-a & 0.478 & 0.628 & \hl{0.130} \\
\bottomrule
\end{tabularx}
\end{table}

Table~\ref{tab:benchmark} compares the predictive accuracy of the proposed method against the benchmark models. It was observed from the table that the proposed approach generally outperforms the benchmark models. For TSS, the proposed model improves the \hl{site-mean} Log-$R^2$ from 0.706 to 0.729, representing a \hl{3.3\%} increase over the second-best DL-RS \hl{and QAA models}, while reducing the Log-MAE from \hl{0.320} to 0.264, corresponding to a \hl{17.5\%} reduction. For DOC, it achieves a \hl{6.4\%} higher \hl{site-mean} Log-$R^2$ (\hl{0.450} vs \hl{0.423}) and a 2.7\% lower Log-MAE (0.390 vs 0.401) than the second-best DL-RS model. A higher improvement was observed for TChl-a, where the proposed method outperforms HyperEST by increasing the \hl{site-mean} Log-$R^2$ from \hl{0.376} to \hl{0.429} (14.1\% improvement) and reducing the Log-MAE from 0.601 to 0.512 (14.8\% reduction). \hl{The per-site benchmark performance is provided in Supplementary Material.}

\begin{table}[!t]
\centering
\caption{Accuracy comparison between the proposed approach and benchmark models. The best and second-best results are shown in bold and underlined, respectively.}
\label{tab:benchmark}
\small
\setlength{\tabcolsep}{3pt}
\begin{tabularx}{\columnwidth}{
    >{\raggedright\arraybackslash}p{1.35cm}
    >{\raggedright\arraybackslash}X
    >{\centering\arraybackslash}p{2.4cm}
    >{\centering\arraybackslash}p{1.35cm}
}
\toprule
Parameter & Model & \hl{Site-mean Log-$R^2$} & Log-MAE \\
\midrule
TSS & Empirical TSS \cite{choo2022spatial} & \hl{0.685} & 0.525 \\
 & \hl{QAA \cite{Lee2002,IOCCG2015QAAv6}}  & \underline{\hl{0.706}} & \underline{\hl{0.320}} \\
 & \hl{MLP} & \hl{0.695} & \hl{0.473} \\
 & DL-RS \cite{unnithan2025} & \underline{\hl{0.706}} & 0.339 \\
 & HyperEST \cite{LUO2025104761} & \hl{0.694} & 0.458 \\
 & Proposed & \textbf{\hl{0.729}} & \textbf{0.264} \\
\midrule
DOC & Empirical DOC \cite{cherukuru2016estimating} & \hl{0.280} & 0.545 \\
 & \hl{MLP} & \hl{0.302} & \hl{0.511} \\
 & DL-RS \cite{unnithan2025} & \underline{\hl{0.423}} & \underline{0.401} \\
 & Proposed & \textbf{\hl{0.450}} & \textbf{0.390} \\
\midrule
TChl-a & Empirical TChl-a \cite{cherukuru2019bio} & \hl{0.176} & 0.787 \\
 & MLP & \hl{0.196} & 0.742 \\
 & HyperEST \cite{LUO2025104761} & \underline{\hl{0.376}} & \underline{0.601} \\
 & Proposed & \textbf{\hl{0.429}} & \textbf{0.512} \\
\bottomrule
\end{tabularx}
\end{table}


The ablation study results are shown in Table~\ref{tab:ablation} where we tested the effects of \hl{(1) Random initialisation: removing the MAML pretraining stage with random initialisation for the model, (2) Supervised pretraining: replacing the MAML pretraining with standard supervised pretraining, and (3) No adaptation: removing the region-specific adaptation stage by applying the base model directly to regional data. Replacing MAML pretraining with random initialisation reduced site-mean Log-$R^2$ from 0.729 to 0.703 for TSS, from 0.450 to 0.391 for DOC, and from 0.429 to 0.360 for TChl-a, while increasing the corresponding Log-MAE values from 0.264, 0.390, and 0.512 to 0.355, 0.448, and 0.607, respectively. Standard supervised pretraining improved performance relative to random initialisation but remained less effective than MAML pretraining, yielding site-mean Log-$R^2$ values of 0.711, 0.421, and 0.392 and Log-MAE values of 0.327, 0.412, and 0.560 for TSS, DOC, and TChl-a, respectively. Removing region-specific adaptation also resulted in a deterioration in performance: site-mean Log-$R^2$ decreased to 0.523, 0.088, and 0.142 for TSS, DOC, and TChl-a, respectively, while Log-MAE increased to 1.055, 0.650, and 0.942.}

\begin{table*}[!t]
\centering
\caption{Ablation study results for the proposed approach.}
\label{tab:ablation}
\setlength{\tabcolsep}{3pt}
\begin{tabularx}{\textwidth}{
    >{\raggedright\arraybackslash}X
    *{3}{
        >{\centering\arraybackslash}p{2.3cm}
        >{\centering\arraybackslash}p{2cm}
    }
}
\toprule
Model Setup & \multicolumn{2}{c}{TSS} & \multicolumn{2}{c}{DOC} & \multicolumn{2}{c}{TChl-a} \\
\cmidrule(lr){2-3}\cmidrule(lr){4-5}\cmidrule(lr){6-7}
 & \hl{Site-mean} Log-$R^2$ & Log-MAE & \hl{Site-mean} Log-$R^2$ & Log-MAE & \hl{Site-mean} Log-$R^2$ & Log-MAE \\
\midrule
Full & \hl{0.729} & 0.264 & \hl{0.450} & 0.390 & \hl{0.429} & 0.512 \\
\hl{Random initialisation} & \hl{0.703} & 0.355 & \hl{0.391} & 0.448 & \hl{0.360} & 0.607 \\
\hl{Supervised pretraining} & \hl{0.711} & \hl{0.327} & \hl{0.421} & \hl{0.412} & \hl{0.392} & \hl{0.560} \\
No adaptation & \hl{0.523} & 1.055 & \hl{0.088} & 0.650 & \hl{0.142} & 0.942 \\
\bottomrule
\end{tabularx}
\end{table*}

\hl{To further assess adaptation to unseen sites and local sample requirements, we conducted a leave-one-site-out evaluation with local adaptation. For each target site, spectral library samples from its geographic region were excluded before generating synthetic data and pretraining the model. We used the same network architecture and synthetic dataset to compare the proposed MAML pretraining and supervised pretraining. Both models were then adapted using 5, 20, or 50 local support samples under identical adaptation settings and evaluated on the same held-out query samples using the temporal block-based cross-validation scheme. Table~\ref{tab:local_sample_efficiency} reports the equally weighted site-mean Log-MAE, where the proposed approach achieved lower errors than supervised pretraining. With 5 local samples, it reduced Log-MAE from 1.129 to 0.792 for TSS, from 0.763 to 0.613 for DOC, and from 1.144 to 0.917 for TChl-a. Errors decreased for both methods as more local samples became available. With 20 local samples, the proposed approach achieved Log-MAE values of 0.676, 0.544, and 0.771 for TSS, DOC, and TChl-a, respectively, which were lower than those obtained by supervised pretraining with 50 samples (0.754, 0.621, and 0.862).}

\begin{table}[!t]
\centering
\caption{\protect\hl{Leave-one-site-out retrieval Log-MAE with different local sample sizes.}}
\label{tab:local_sample_efficiency}
\small
\setlength{\tabcolsep}{2pt}
\renewcommand{\arraystretch}{1.15}

\begin{tabular*}{\columnwidth}{
@{\extracolsep{\fill}}
>{\centering\arraybackslash}p{0.12\columnwidth}
>{\raggedright\arraybackslash}p{0.34\columnwidth}
*{3}{>{\centering\arraybackslash}p{0.12\columnwidth}}
@{}
}
\toprule
\hl{Local samples}
& \hl{Method} & \hl{TSS} & \hl{DOC} & \hl{TChl-a} \\
\midrule
\hl{5} & \hl{Supervised Pretraining} & \hl{1.129} & \hl{0.763} & \hl{1.144} \\
\hl{5} & \hl{Proposed} & \hl{0.792} & \hl{0.613} & \hl{0.917} \\
\addlinespace
\hl{20} & \hl{Supervised Pretraining} & \hl{0.943} & \hl{0.692} & \hl{0.953} \\
\hl{20} & \hl{Proposed} & \hl{0.676} & \hl{0.544} & \hl{0.771} \\
\addlinespace
\hl{50} & \hl{Supervised Pretraining} & \hl{0.754} & \hl{0.621} & \hl{0.862} \\
\hl{50} & \hl{Proposed} & \hl{0.627} & \hl{0.512} & \hl{0.689} \\
\bottomrule
\end{tabular*}
\end{table}

\subsection{Time-series analysis of \hl{WQ} and $R_{rs}$ data}
\label{ssec:timeseries_analysis}

\begin{figure*}[!t]
\centering
\includegraphics[width=0.85\textwidth]{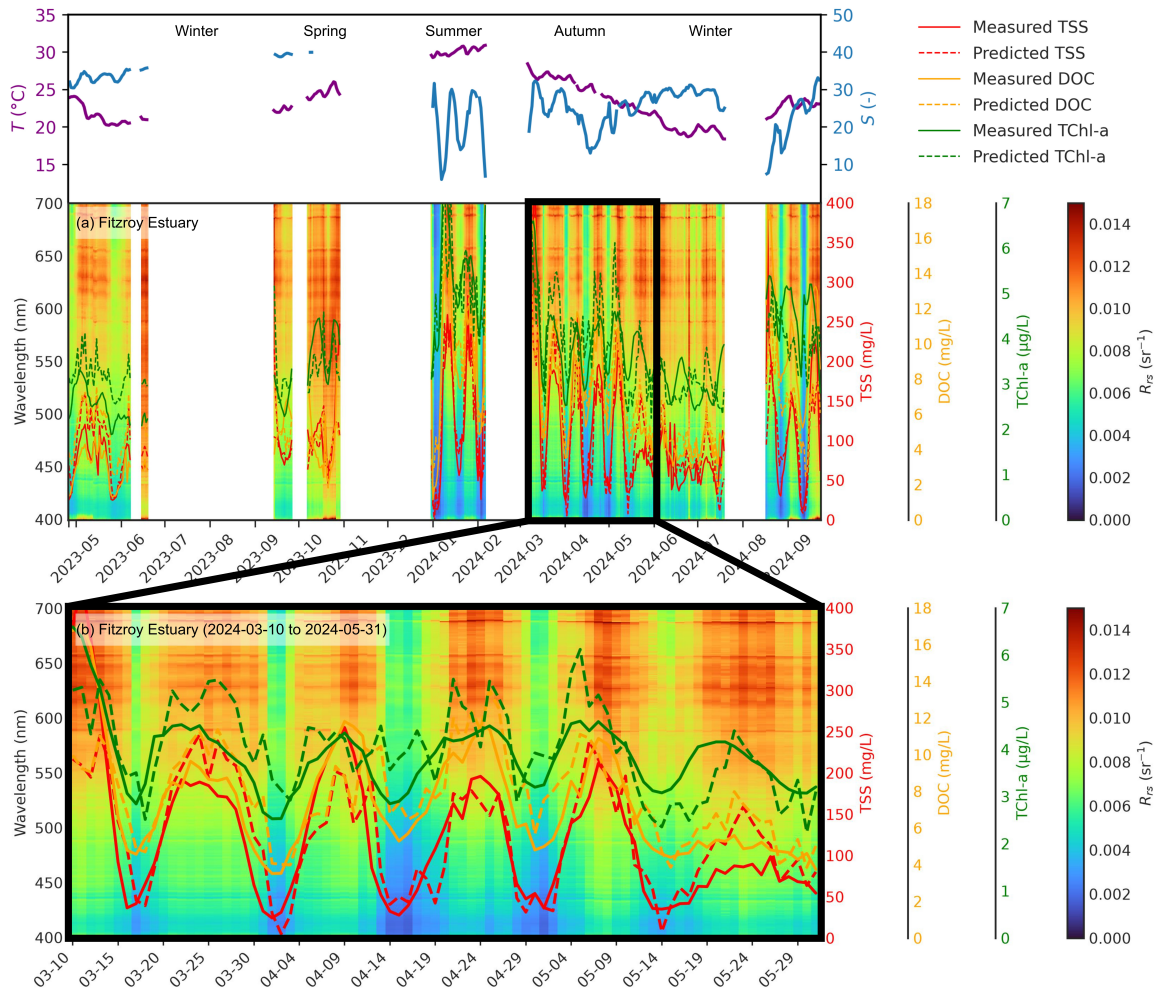}
\caption{(a) Temporal variation of water-leaving $R_{rs}$ and corresponding in situ measurements of temperature ($T$), salinity ($S$), TSS, DOC, and TChl-a at the experimental site of Fitzroy Estuary over the experimental period from 26 April 2023 to 23 September 2024. The TSS, DOC, and TChl-a predictions generated by the proposed approach from hyperspectral $R_{rs}$ observations in the test folds are also shown. (b) Zoomed-in view for the period between 10 March and 31 May 2024. Background colour shows hyperspectral $R_{rs}$ as a function of wavelength (400--700 nm) and time. Blank regions in the plots denote periods of instrument malfunction. \label{fig:ts_fe}}
\end{figure*}

Fig. \ref{fig:ts_fe} shows time-series variations in hyperspectral $R_{rs}$ and corresponding \hl{WQ} parameters (TSS, DOC, and TChl-a) and ancillary physical variables (temperature and salinity) at the experimental site of Fitzroy Estuary, with Fig. \ref{fig:ts_fe}a showing data over the entire experimental period from 26 April 2023 to 23 September 2024, and Fig. \ref{fig:ts_fe}b showing the zoomed-in view for the period between 10 March and 31 May 2024. It was observed from the figure that, the Fitzroy Estuary site, characterised by high river influence and a subtropical climate (Table~\ref{tab:sites}), exhibits high TSS concentrations with a periodic variability over time, alongside elevated and highly dynamic levels of DOC and TChl-a. The periodic variability likely reflects the combined influence of tidal cycles and fluctuations in Fitzroy River discharge, as evidenced by the salinity variations shown in the figure, which indicate varying degrees of mixing between river water and seawater.

The time series of $R_{rs}$ at the Fitzroy Estuary site (Fig.~\ref{fig:ts_fe}) shows relatively high $R_{rs}$ values, reaching up to approximately 0.06 sr\textsuperscript{\textminus 1}, particularly in the green to red spectral region (550--650 nm). The time-series $R_{rs}$ measurements also exhibit pronounced temporal variability, including a periodic pattern over the observation period (Fig.~\ref{fig:ts_fe}). Given that the Fitzroy Estuary is a river-influenced system characterised by relatively high TSS concentrations, the elevated $R_{rs}$ in the green-to-red region is likely driven by enhanced particulate backscattering associated with high sediment loads. The observed periodic variability in $R_{rs}$ may further reflect temporal changes in \hl{WQ} properties, influenced by tidal dynamics and variations in river discharge.

\begin{figure*}[!t]
\centering
\includegraphics[width=0.77\textwidth]{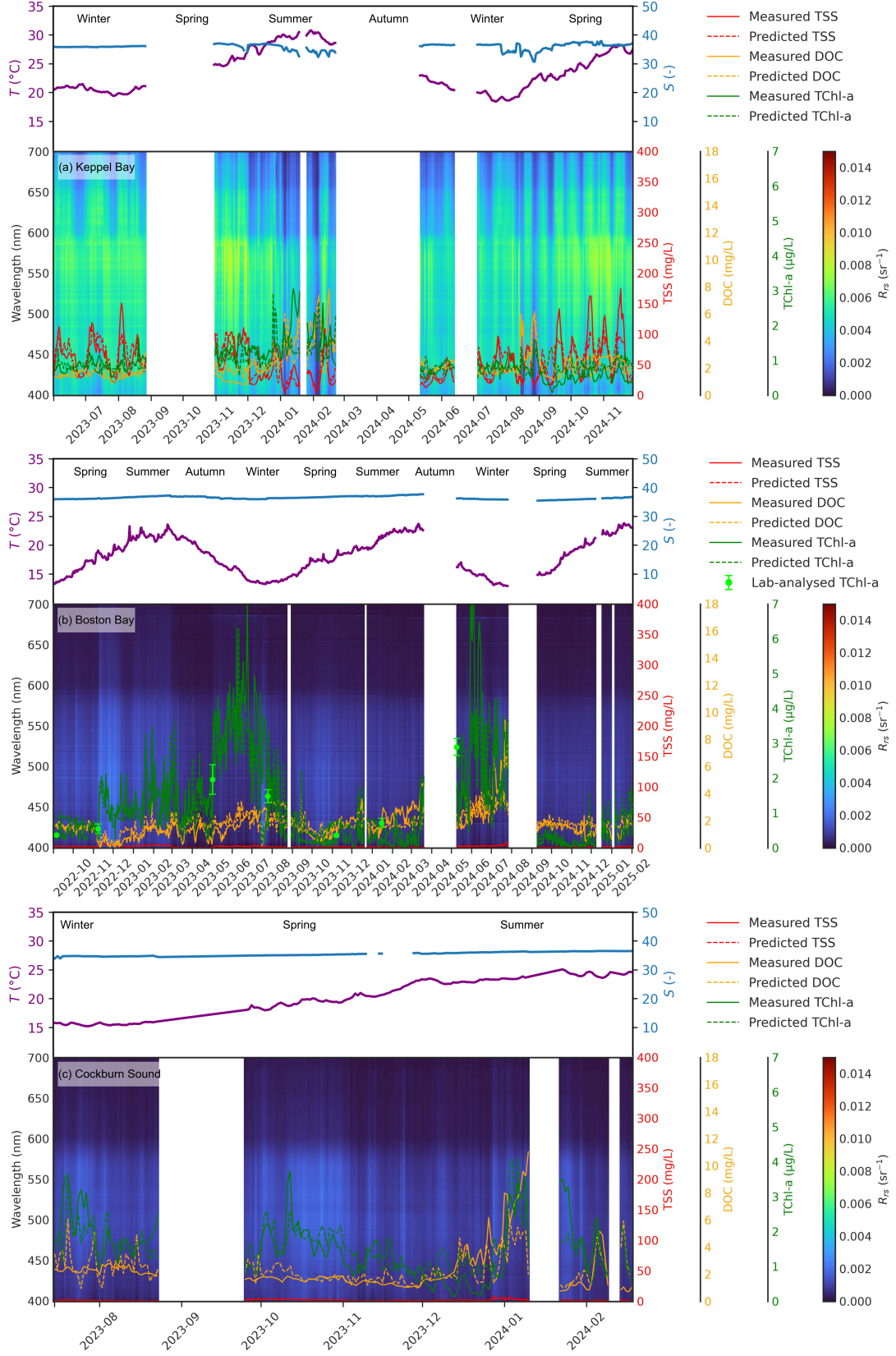}
\caption{Temporal variation of water-leaving $R_{rs}$ and corresponding in situ measurements of temperature ($T$), salinity ($S$), TSS, DOC, and TChl-a at the experimental sites of (a) Keppel Bay, (b) Boston Bay, and (c) Cockburn Sound. Background colour shows hyperspectral $R_{rs}$ as a function of wavelength (400--700 nm) and time. The TSS, DOC, and TChl-a predictions generated by the proposed approach from hyperspectral $R_{rs}$ observations in the test folds are also shown. Laboratory-analysed grab samples of TChl-a (mean $\pm$ standard deviation) are shown as light green markers. Blank regions in the plots denote periods of instrument malfunction. \label{fig:ts_sites} }
\end{figure*}

The time-series plots for other sites (Keppel Bay, Boston Bay, and Cockburn Sound) are shown in Fig. \ref{fig:ts_sites}. It is worth noting that the Lucinda Jetty result is not shown in the figure because the data were not sampled continuously in time at this site. Fig. \ref{fig:ts_sites}a shows time-series variations in $R_{rs}$, TSS, DOC, TChl-a, temperature, and salinity at the Keppel Bay site over the experimental period from 1 June 2023 to 28 November 2024. This site is located downstream of the Fitzroy Estuary site and is closer to the open ocean. Accordingly, higher salinity was observed at Keppel Bay compared with the Fitzroy Estuary site. For the \hl{WQ} parameters, concentrations of TSS, DOC, and TChl-a at the Keppel Bay site are generally lower than those at the Fitzroy Estuary site. Periodic variations are also evident in the time-series \hl{WQ} measurements at the Keppel Bay site, although they are less pronounced than those observed at the Fitzroy Estuary site. As compared with the Fitzroy Estuary site, lower levels of $R_{rs}$ were observed at the Keppel Bay site, as shown in Fig. \ref{fig:ts_sites}a. The peak $R_{rs}$ is generally below 0.04 sr\textsuperscript{\textminus 1} and centred in the green spectral region around 550--600 nm.

Fig. \ref{fig:ts_sites}b shows the time-series data for the Boston Bay site over the experimental period from 1 September 2022 to 2 February 2025. Different from the subtropical Fitzroy Estuary and Keppel Bay sites, the Boston Bay site is a temperate site without direct river influence (Table~\ref{tab:sites}). Accordingly, it was observed from Fig. \ref{fig:ts_sites}b that the Boston Bay site is characterised by consistently low TSS and DOC and minimal temporal variability, indicative of clearer and more optically stable waters. The $R_{rs}$ levels at the Boston Bay site are generally lower than the Fitzroy Estuary and Keppel Bay sites, with the peak $R_{rs}$ below 0.02 sr\textsuperscript{\textminus 1} and centred in the green/blue spectral region of 450--550 nm. 

An important observation from Fig. \ref{fig:ts_sites}b is the clear seasonal pattern in TChl-a concentrations at the Boston Bay site, which increase during the winter months. This seasonal rise begins around March/April each year over the experimental period, \hl{remains elevated into winter before declining around July/August}. To verify this pattern, grab samples were collected at the Boston Bay site on several occasions during the experiment, and TChl-a concentrations were analysed by the CSIRO Hydrochemistry Laboratories in Hobart, Tasmania. As shown in Fig. \ref{fig:ts_sites}b, the laboratory-analysed TChl-a values are in good agreement with the in situ time-series measurements, consistently showing higher concentrations between March/April and July/August. \hl{In March 2025, shortly after the Boston Bay observation period ended,} a harmful algal bloom broke out along the South Australian coast\footnote{\url{https://en.wikipedia.org/wiki/2025_algal_bloom_in_South_Australia}}, including the Boston Bay region, resulting in mass fish deaths and significant impacts on marine ecosystems, fisheries, aquaculture, tourism, and coastal activities. The observation reported in this study may therefore provide important baseline information on seasonal phytoplankton dynamics and could contribute to a better understanding of environmental conditions that precede or favour the development of such bloom events.

Fig. \ref{fig:ts_sites}c shows the time-series data for the Cockburn Sound site over the experimental period from 15 July 2023 to 18 February 2024. As observed from the figure, Cockburn Sound, being a temperate embayment with low freshwater influence (Table~\ref{tab:sites}), exhibits generally low to moderate concentrations of TSS, DOC, and TChl-a over the experimental period, with some temporal variability. Correspondingly, this site displays relatively low $R_{rs}$ values, with peak $R_{rs}$ typically below 0.015 sr\textsuperscript{\textminus 1} and centred in the green spectral region around 500--550 nm.

It was also observed from Figs. \ref{fig:ts_fe} and \ref{fig:ts_sites} that, despite clear regional differences among the experimental sites, the \hl{WQ} parameters predicted from hyperspectral $R_{rs}$ observations using the proposed method generally agree well with the in situ \hl{WQ} measurements in both magnitude and temporal dynamics. This consistency demonstrates the capability of in situ hyperspectral sensing to reliably capture \hl{WQ} variability over time, showing its potential as a cost-effective approach for continuous time-series monitoring of WQ.

\subsection{\hl{Sensitivity to Additional $R_{rs}$ Noise}}
\label{ssec:noise-sensitivity}

\begin{figure*}[!t]
    \centering
    \includegraphics[width=\textwidth]{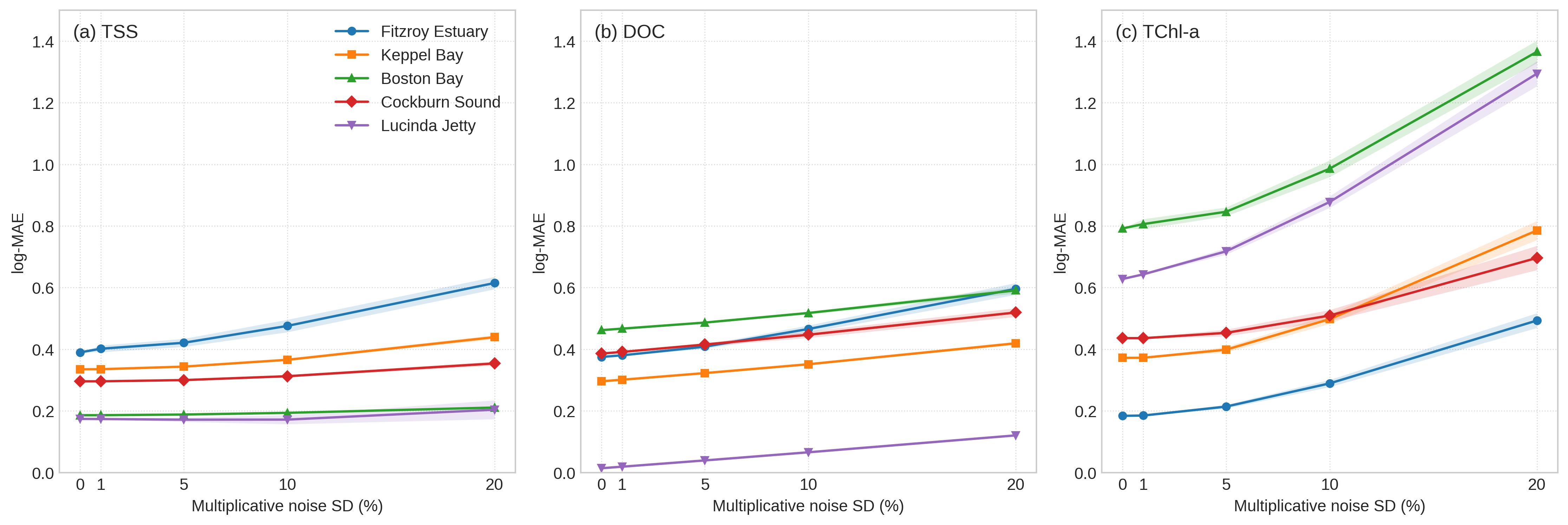}
    \caption[Noise sensitivity of water-quality retrievals.]{\hl{Sensitivity of TSS, DOC, and TChl-a retrieval to additional independent bandwise multiplicative Gaussian noise. Lines show mean site-specific Log-MAE, and shaded bands indicate $\pm$ one standard deviation across ten noise realizations at each noise level.}}
    \label{fig:noise-sensitivity}
\end{figure*}

\hl{To assess sensitivity to additional $R_{rs}$ noise, we added independent bandwise multiplicative Gaussian noise to held-out query spectra as $\widetilde{R}_{rs}=R_{rs}
(1+\sigma\epsilon)$, where $\epsilon\sim\mathcal{N}(0,1)$ and $\sigma=0\%$, 1\%, 5\%, 10\%, or 20\%. Fig.~\ref{fig:noise-sensitivity} summarises ten realisations per nonzero noise level, with shaded bands indicating one standard deviation around the mean. At 20\% noise, the Log-MAE increased from
0.264 to 0.353 for TSS, from 0.390 to 0.532 for DOC, and from 0.512 to 0.957 for TChl-a, respectively.}

\hl{Spectral denoising may help reduce random fluctuations in $R_{rs}$, potentially improving retrieval stability. However, excessive smoothing may alter absorption features and spectral shapes relevant to WQ retrieval. The choice and strength of denoising therefore require a balance between noise suppression and preservation of useful spectral information. Future work should compare retrieval performance with and without an additional denoising step to assess whether denoising improves accuracy across parameters and sites.}

\subsection{\hl{Application to PACE OCI Observations}}
\label{ssec:pace-application}

\begin{figure*}[!t]
    \centering
    \includegraphics[width=\textwidth]{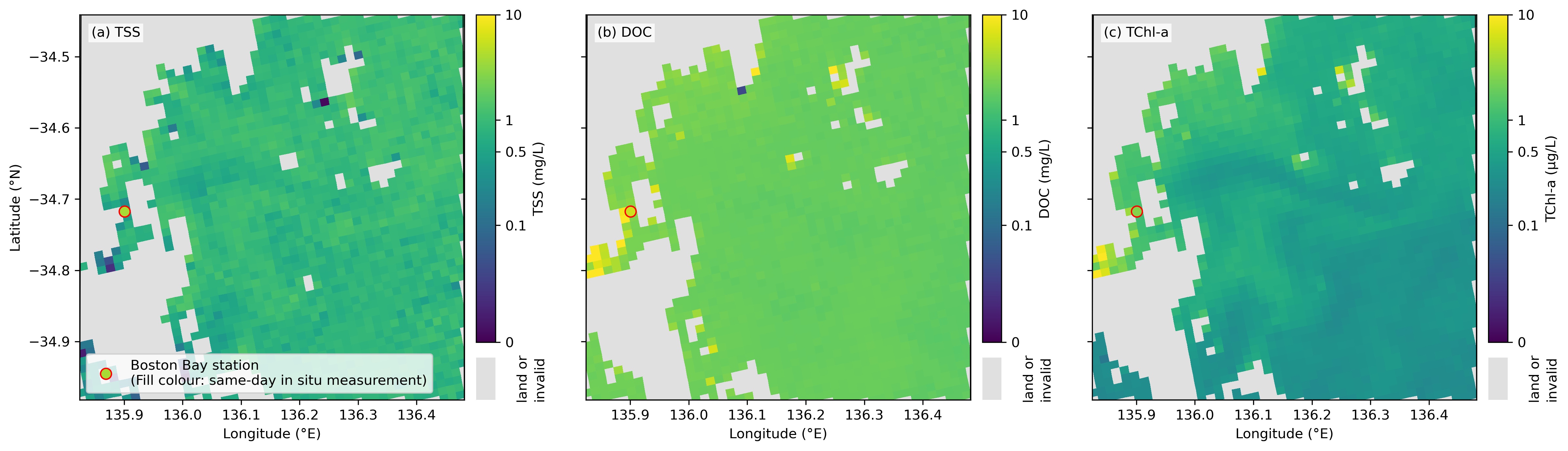}
    \caption[Application to PACE OCI observations.]{\hl{Estimated (a) TSS, (b) DOC, and (c) TChl-a from PACE OCI observations acquired on 6 July 2024 using the Boston Bay-adapted model. The red-outlined circle marks the Boston Bay station; its fill colour represents the same-day in situ concentration on the corresponding colour scale. Grey areas indicate land or invalid pixels.}}
    \label{fig:pace-application}
\end{figure*}

\hl{We applied the Boston Bay-adapted model to PACE OCI observations acquired on 6 July 2024 over a $60\times60$~km region encompassing Boston Bay (Fig.~\mbox{\ref{fig:pace-application}}). At the Boston Bay station, satellite-derived TSS, DOC, and TChl-a were 3.511~mg/L, 6.232~mg/L, and 2.051~$\mu$g/L, respectively, compared with same-day in situ values of 3.870~mg/L, 5.806~mg/L, and 2.655~$\mu$g/L. This example demonstrates the potential of applying the proposed approach to satellite observations, although further validation using independent satellite--in situ matchups across multiple scenes and regions is needed in future work. Future work should also assess systematic and spectrally correlated atmospheric-correction errors, which are not represented by the independent Gaussian perturbations tested here.}

\subsection{\hl{Comparison with existing methods}}
\label{ssec:comparison_with_existing_methods}

\hl{
Traditional empirical methods retrieve WQ parameters using locally calibrated relationships between $R_{rs}$ and target variables, including the TSS model by Choo et al.~\cite{choo2022spatial}, the DOC model by Cherukuru et al.~\cite{cherukuru2016estimating}, and the TChl-a model by Cherukuru et al.~\cite{cherukuru2019bio}. Although computationally efficient and interpretable, their transferability can be limited by regional differences in constituent composition and bio-optical properties, often requiring recalibration.

Recent machine-learning approaches have advanced multi-parameter retrieval. Pahlevan et al.~\cite{pahlevan2022simultaneous} developed mixture density networks for simultaneous retrieval of chlorophyll-a,
TSS, and CDOM absorption from multiple satellite sensors. O'Shea et al.~\cite{OShea2023} developed probabilistic hyperspectral retrieval of absorbing IOPs and WQ parameters. Werther et al.~\cite{werther2025generalization} investigated probabilistic neural networks for hyperspectral retrieval of absorption properties. These studies establish simultaneous retrieval as an existing capability. Our contribution instead emphasises preparing a shared model initialisation for adaptation to regional bio-optical conditions.

Pretraining also has established precedent. Unnithan et al.~\cite{unnithan2025} developed DL-RS using an augmented
spectral library and bio-optical forward modelling, whereas HyperEST~\cite{LUO2025104761} learns representations from unlabelled spectra through self-supervised pretraining. Our framework organises synthetic samples into SIOP-based tasks and optimises query-set performance after support-set adaptation. Unlike conventional supervised pretraining followed by fine-tuning, MAML explicitly includes post-adaptation performance in its pretraining objective.
The additional capability is learning an initialisation prepared for regional adaptation, rather than transferring parameters optimised solely for source-data prediction.

There have been attempts to incorporate physical knowledge in data-driven retrieval of WQ parameters from spectral reflectance data. For example, OCRT-Net by Cao et al.~\cite{cao2026ocrt} incorporates physical knowledge by embedding ocean-colour radiative-transfer theory within its architecture. In comparison, our framework incorporates physical knowledge through bio-optical forward modelling to generate synthetic WQ--$R_{rs}$ data, while SIOP similarity guides the construction of meta-learning tasks.

Overall, the additional capability of the present framework is to learn a shared initialisation explicitly optimised for regional adaptation of multi-parameter WQ retrieval. Beyond predicting multiple parameters jointly, it uses SIOP-based synthetic tasks to train the model for performance after task-specific updates. This distinguishes its objective from conventional supervised pretraining followed by fine-tuning. IOPs inform the simulated optical relationships and task construction, while the network directly predicts TSS, DOC, and TChl-a. Physical knowledge is incorporated through synthetic-data generation. Our contribution therefore lies in combining these elements to
construct a retrieval model for region-adaptive hyperspectral retrieval of WQ parameters.
}

\subsection{\hl{Limitation and future work}}
\label{ssec: limitations_and_future_work}

\hl{
It is worth noting that the DOC retrieval has a mechanistic limitation: visible reflectance is sensitive to CDOM absorption, whereas the non-chromophoric fraction of DOC is not directly detectable within the 400--700 nm range used here. Consequently, the DOC is estimated indirectly through regional covariation between DOC and CDOM. This relationship may vary with organic matter sources, mixing, and biogeochemical processing. Accordingly, $a_y^*(\lambda)=a_y(\lambda)/\mathrm{DOC}$ is an empirical DOC-normalised CDOM absorption coefficient. Changes in this ratio can weaken the correspondence between optical observations and DOC concentration. Regional adaptation may help accommodate differences in the DOC--CDOM relationship represented by local training samples, but cannot make non-chromophoric DOC directly observable. Transferability may therefore be limited when this relationship differs substantially from those represented in the spectral library and adaptation data, or changes over time. Future work should expand paired DOC and CDOM absorption measurements across regions and seasons, evaluate performance under such changes, and investigate whether ancillary environmental information can improve DOC estimation.
}

\section{Conclusion}
\label{sec:conclusion}

This study presented a new framework for retrieving key \hl{WQ} parameters, including TSS, DOC, and TChl-a, from in situ hyperspectral $R_{rs}$ measurements. The proposed method integrates physical knowledge with a data-driven learning strategy through a two-stage physics-aware meta-learning framework. In the first stage, a physics-based bio-optical forward model was used to generate a large physics-guided synthetic dataset from a continental bio-optical spectral library. This dataset was then used to pretrain a region-agnostic base model with meta-learning, allowing the model to learn fundamental physical relationships. In the second stage, the pretrained base model was adapted to individual regions using local in situ observations, allowing the framework to account for region-specific bio-optical variability.

Evaluation across five experimental sites showed clear regional differences in both \hl{WQ} parameters and hyperspectral $R_{rs}$ signatures. \hl{Evaluation results suggested that} the proposed approach outperformed \hl{the} benchmark models. Relative to the second-best model, it improved \hl{the site-mean} Log-$R^2$ by \hl{3.3\%}, \hl{6.4\%}, and 14.1\% for TSS, DOC, and TChl-a, respectively, while reducing Log-MAE by \hl{17.5\%}, 2.7\%, and 14.8\%. The ablation results showed that both physics-aware pretraining and region-specific adaptation contributed to retrieval accuracy. The predicted time series also agreed well with in situ measurements in both magnitude and temporal variation, showing the potential of the framework for continuous coastal \hl{WQ} monitoring.

While the results are promising, there remain opportunities for further improvement. The current evaluation focuses on a limited number of coastal and estuarine sites, which may not fully represent the broader diversity of optical water types, particularly inland waters with stronger spatiotemporal variability. Future work would be on extending validation to additional aquatic environments, and exploring the application of the proposed framework to satellite-based hyperspectral observations.
\section*{Acknowledgement}

\hl{The authors would like to thank the anonymous Associate Editor and Reviewers for their insightful comments and constructive suggestions, which significantly improved the quality and clarity of this manuscript.} We would like to acknowledge the in situ data from Lucinda Jetty Coastal Observatory (Principal Investigator: Dr Thomas Schroeder). These data were sourced from Australia’s Integrated Marine Observing System (IMOS). IMOS is enabled by the National Collaborative Research Infrastructure Strategy (NCRIS). It is operated by a consortium of institutions as an unincorporated joint venture, with the University of Tasmania as Lead Agent. We acknowledge the SARDI field team (Mr Ian Moody and Mr Paul Malthouse) for their help with Boston Bay buoy setup and maintenance. We are grateful to Dr Albertina Dias with CSIRO Hydrochemistry Laboratories, Hobart, Tasmania, for analysing the grab water samples.

\hl{The code associated with this study is available at}
\url{https://doi.org/10.5281/zenodo.23786563}.
\hl{Publicly available datasets include the Fitzroy River and Keppel Bay QLD Bio-optics Sample Collections at}
\url{https://doi.org/10.25919/w2jj-ps76};
\hl{the NSW Coastal Outflows Bio-optics collections at}
\url{https://doi.org/10.25919/z7xf-z004},
\url{https://doi.org/10.25919/1jsj-j657},
\hl{and}
\url{https://doi.org/10.25919/h85w-1v75};
\hl{and the Moreton Bay 2011 post-flood bio-optical and phytoplankton dataset at}
\url{https://doi.org/10.25919/gf5w-9711}.
\hl{The CoastColour Round Robin datasets are available through PANGAEA at}
\url{https://doi.org/10.1594/PANGAEA.841950}.
\hl{Additional bio-optical observations are accessible through the IMOS Bio Optical Database of Australian Waters via the AODN portal at}
\url{https://portal.aodn.org.au/search?uuid=97b9fe73-ee44-437f-b2ae-5b8613f81042}.
\hl{The Lucinda Jetty hyperspectral reflectance observations are available through the IMOS/AODN HyperOCR archive at}
\url{https://thredds.aodn.org.au/thredds/catalog/IMOS/SRS/OC/LJCO/HyperOCR-hourly/catalog.html}.
\hl{Some research datasets used in this study are not publicly available due to intellectual property restrictions.}

\bibliographystyle{IEEEtran}
\bibliography{ref}

\begin{IEEEbiography}[{\includegraphics[width=1in,height=1.25in,clip,keepaspectratio]{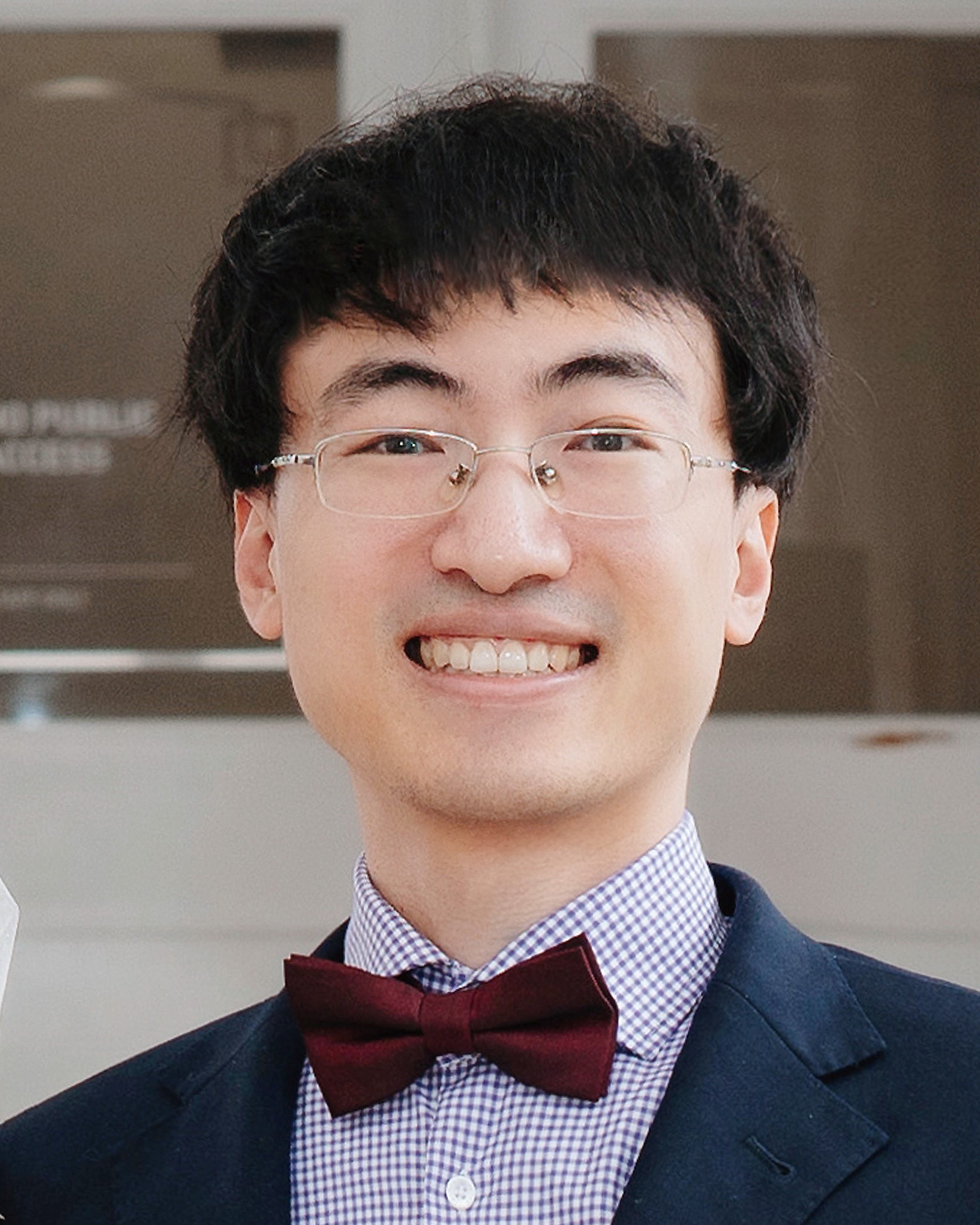}}]{Yiqing Guo}
received the B.Eng.\ and M.Eng.\ degrees from Beihang University, Beijing, China, in 2012 and 2015, respectively, and the Ph.D.\ degree from the University of New South Wales, Canberra, Australia, in 2019. He joined the Commonwealth Scientific and Industrial Research Organisation (CSIRO), Australia, in 2020, where he is currently a Research Scientist. His research interests include remote sensing and machine learning, and their applications to environmental problems.
\end{IEEEbiography}

\begin{IEEEbiography}[{\includegraphics[width=1in,height=1.25in,clip,keepaspectratio]{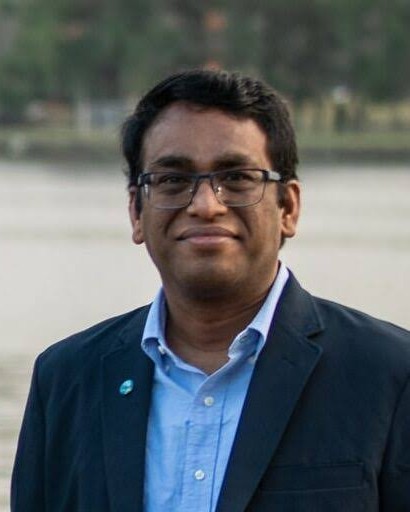}}]{Nagur R. C. Cherukuru}
received the M.Sc.\ degree in Geophysics from Andhra University, India, in 1998, and the Ph.D.\ degree in Marine Optics and Remote Sensing from the University of Plymouth, Plymouth, U.K., in 2005. He joined the Commonwealth Scientific and Industrial Research Organisation (CSIRO), Australia, in 2006, where he is currently a Principal Research Scientist. His research interests include bio-optical modelling and remote sensing of water quality.
\end{IEEEbiography}

\begin{IEEEbiography}[{\includegraphics[width=1in,height=1.25in,clip,keepaspectratio]{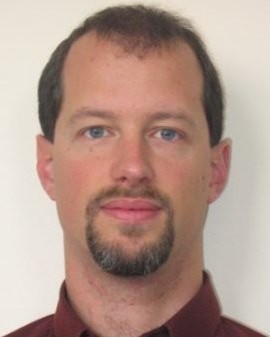}}]{Eric A. Lehmann}
received the Dipl.\ El.-Ing.\ degree from the Swiss Federal Institute of Technology (ETH), Zurich, Switzerland, in 1999, and the M.Phil.\ and Ph.D.\ degrees from the Australian National University, Canberra, ACT, Australia, in 2000 and 2004, respectively. He was a Research Fellow with Western Australian Telecommunications Research Institute from 2005 to 2008. He joined the Commonwealth Scientific and Industrial Research Organisation (CSIRO), Australia, in 2008, where he is currently a Senior Research Scientist. His research interests include remote sensing and environmental monitoring.
\end{IEEEbiography}

\begin{IEEEbiography}[{\includegraphics[width=1in,height=1.25in,clip,keepaspectratio]{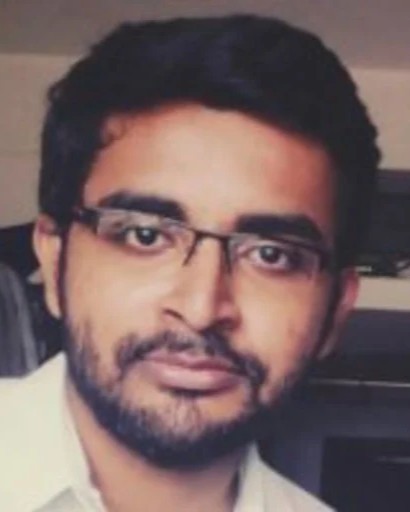}}]{S. L. Kesav Unnithan}
received the B.Tech.\ degree from Cochin University of Science and Technology, Kochi, India, in 2016, the M.Tech.\ degree from the Indian Institute of Space Science and Technology, Thiruvananthapuram, India, in 2018, and the Ph.D.\ degree from the Indian Institute of Technology Bombay, Mumbai, India, and Monash University, Melbourne, VIC, Australia, in 2023. He joined the Commonwealth Scientific and Industrial Research Organisation (CSIRO), Australia, in 2023, where he is currently a Research Scientist. His research interests include remote sensing, geospatial artificial intelligence, and water quality monitoring.
\end{IEEEbiography}

\begin{IEEEbiography}[{\includegraphics[width=1in,height=1.25in,clip,keepaspectratio]{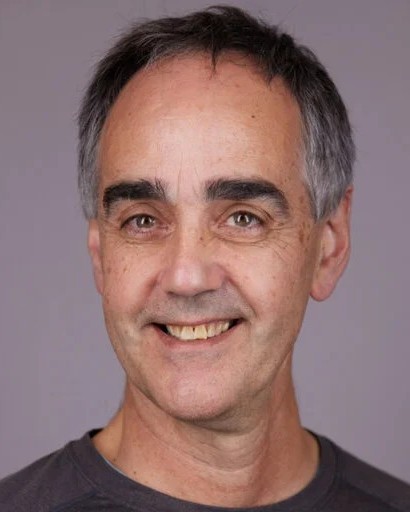}}]{Tim J. Malthus}
received the B.Sc., Dip.Sci., and Ph.D. degrees from the University of Otago, Dunedin, New Zealand, in 1978, 1980, and 1986, respectively. He was a Senior Lecturer with the University of Edinburgh, Edinburgh, U.K. from 1994 to 2009. In 2009, he joined the Commonwealth Scientific and Industrial Research Organisation (CSIRO), Australia, where he is currently a Senior Principal Research Scientist. His research interests include hyperspectral sensing and water quality monitoring.
\end{IEEEbiography}

\begin{IEEEbiography}[{\includegraphics[width=1in,height=1.25in,clip,keepaspectratio]{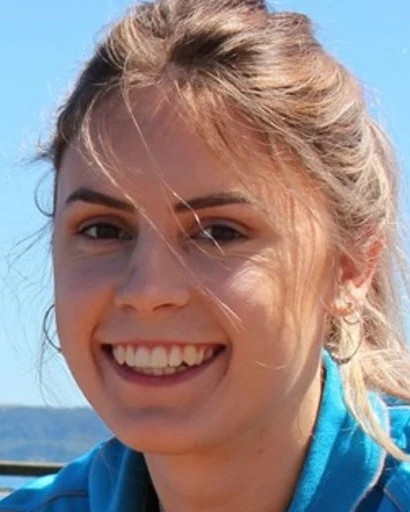}}]{Gemma Kerrisk}
received the B.Sc.\ degree in Genetics from Massey University, Palmerston North, New Zealand, in 2017, and the Postgraduate Diploma in Environmental Management from the University of Auckland, Auckland, New Zealand, in 2018. She joined the Commonwealth Scientific and Industrial Research Organisation (CSIRO), Australia, in 2019, where she is currently an Experimental Scientist. Her research interests include aquatic remote sensing, optical oceanography, and water quality monitoring.
\end{IEEEbiography}

\begin{IEEEbiography}[{\includegraphics[width=1in,height=1.25in,clip,keepaspectratio]{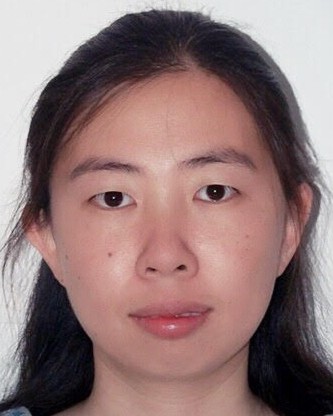}}]{Xiubin Qi}
received the B.Sc.\ and M.Sc.\ degrees from Nanjing University, Nanjing, China, and the Ph.D.\ degree from the University of California, Davis, CA, USA. She joined the Commonwealth Scientific and Industrial Research Organisation (CSIRO), Australia, in 2008, where she is currently a Senior Research Scientist. Her research interests include aquatic sensing technologies, satellite calibration and validation, and water quality monitoring.
\end{IEEEbiography}

\begin{IEEEbiography}[{\includegraphics[width=1in,height=1.25in,clip,keepaspectratio]{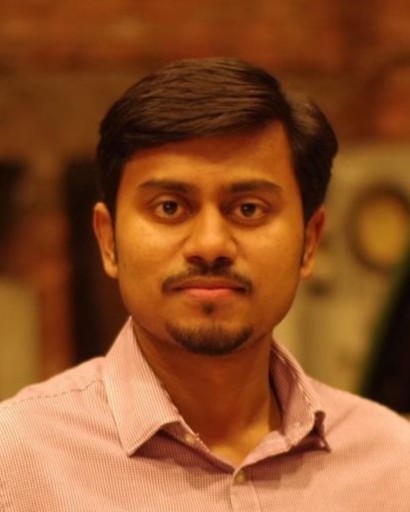}}]{Faisal Islam}
received the B.Tech.\ and M.Tech.\ degrees from the Indian Institute of Technology Delhi, Kanpur, India, in 2014 and 2015, respectively, and the Ph.D.\ degree from Mines Paris -- PSL, Paris, France, in 2020. He worked as a Research Fellow with the University of New South Wales, Sydney, NSW, Australia, before joining the Commonwealth Scientific and Industrial Research Organisation (CSIRO), Australia, in 2023 as a Research Scientist. His research interests include statistical modelling and artificial intelligence for environmental and engineering applications.
\end{IEEEbiography}

\begin{IEEEbiography}[{\includegraphics[width=1in,height=1.25in,clip,keepaspectratio]{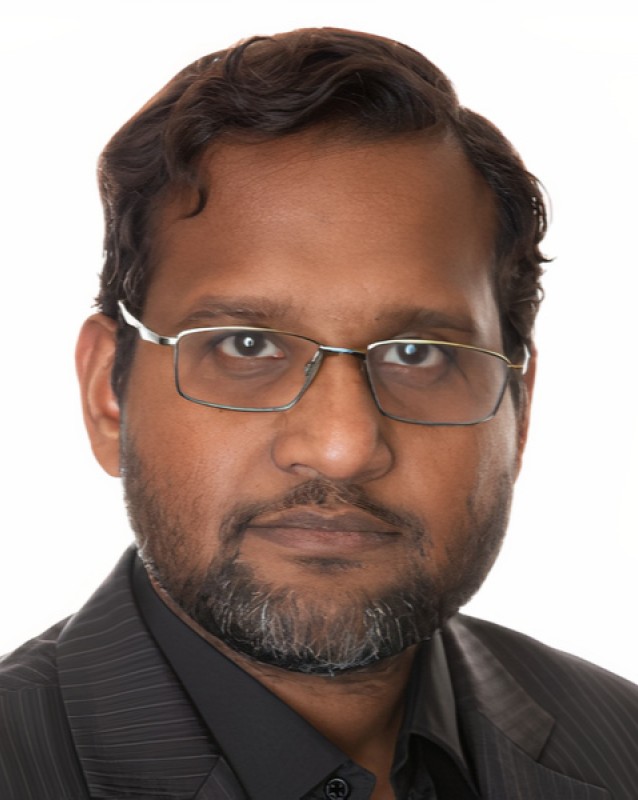}}]{Tisham Dhar}
received the B.Eng.\ degree from the University of Adelaide, Australia, in 2006. He held leadership roles in engineering and software development in government agencies, industry, and startup companies before joining the Commonwealth Scientific and Industrial Research Organisation (CSIRO), Australia, in 2023, where he is currently a Senior Engineer. His research interests include geospatial software, Earth observation data systems, and open-source technologies.
\end{IEEEbiography}

\begin{IEEEbiography}[{\includegraphics[width=1in,height=1.25in,clip,keepaspectratio]{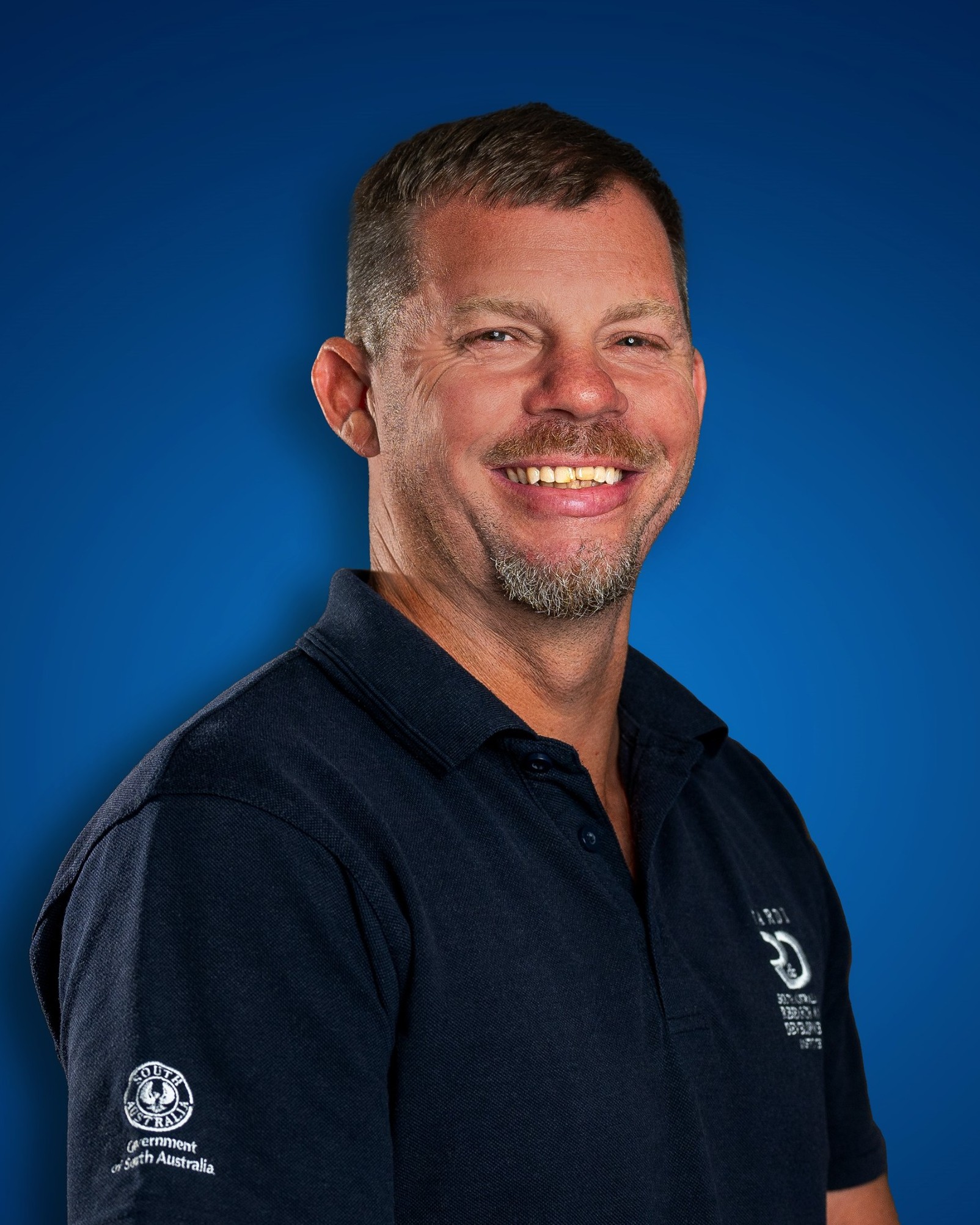}}]{Mark J. Doubell}
received the M.Sc.\ degree in Marine Sciences from Flinders University, Adelaide, SA, Australia,
in 2001 and the Ph.D.\ degree in Biological Oceanography in 2007. He was a Postdoctoral Fellow with
Tokyo University of Marine Science and Technology, Tokyo, Japan, from 2007 to 2009. He joined the South Australian Research and Development Institute
(SARDI), Adelaide, SA, Australia, in 2009, where he is currently a Principal Scientist in Oceanography. His research interests include aquatic remote sensing, ocean microstructure and phytoplankton ecology.

\end{IEEEbiography}

\end{document}